\pdfoutput=1

\documentclass[11pt]{article}
\usepackage{acl}
\usepackage{times}
\usepackage{latexsym}
\usepackage[T1]{fontenc}
\usepackage[utf8]{inputenc}
\usepackage{microtype}

\pdfoutput=1

\usepackage{booktabs}
\usepackage{booktabs}
\usepackage{amsfonts}
\usepackage{nicefrac}
\usepackage{microtype}
\usepackage{todonotes}
\usepackage{listings}
\usepackage{algorithm}
\usepackage{algpseudocode}
\usepackage{soul}
\usepackage{tabularx}
\usepackage{pifont}
\usepackage{xcolor}
\usepackage{graphicx}
\usepackage{subcaption}
\usepackage{hyperref}
\usepackage{comment}
\usepackage{tcolorbox}
\usepackage{nicematrix}
\usepackage{tabu}
\usepackage{colortbl}
\usepackage{xcolor}
\usepackage{makecell}

 \usepackage{array}
 \usepackage{threeparttable}
 \usepackage{xcolor}
 
\usepackage[normalem]{ulem}

\usepackage{hyperref}
\usepackage{multirow}
\usepackage{graphicx}
\usepackage{pdfpages}
\usepackage{url}
\usepackage{xcolor}
\usepackage{epsfig}
\usepackage{adjustbox}
\usepackage{amsfonts}
\usepackage{amsmath}
\usepackage{amssymb}
\usepackage{booktabs} 
\usepackage{comment}
\usepackage{caption}
\usepackage{subcaption}
\usepackage{textcomp}
\usepackage{relsize}
\usepackage{stmaryrd}
\usepackage{bbm}
\usepackage{rotating}
\usepackage{helvet}
\usepackage{courier}
\usepackage{natbib}
\usepackage{cleveref}
\usepackage{xspace}
\usepackage{enumitem}
\usepackage{makecell}

\usepackage{hyperref}
\usepackage{multirow}
\usepackage{graphicx}
\usepackage{pdfpages}
\usepackage{url}
\usepackage{xcolor}
\usepackage{epsfig}
\usepackage{adjustbox}
\usepackage{amsfonts}
\usepackage{amsmath}
\usepackage{amssymb}
\usepackage{booktabs} 
\usepackage{comment}
\usepackage{caption}
\usepackage{subcaption}
\usepackage{textcomp}
\usepackage{relsize}
\usepackage{stmaryrd}
\usepackage{bbm}
\usepackage{rotating}
\usepackage{helvet}
\usepackage{courier}
\usepackage{natbib}
\usepackage{cleveref}
\usepackage{xspace}
\usepackage{enumitem}
\usepackage{makecell}
\PassOptionsToPackage{table}{xcolor}

\newcolumntype{X}{>{\centering\arraybackslash}m{0.17\linewidth}}
\newcolumntype{Y}{>{\centering\arraybackslash}m{0.08\linewidth}}

\usepackage{footmisc}
\DefineFNsymbols{mySymbols}{{\ensuremath\dagger}{\ensuremath\ddagger}\S\P
   *{**}{\ensuremath{\dagger\dagger}}{\ensuremath{\ddagger\ddagger}}}
\setfnsymbol{mySymbols}

\title{Unveiling Implicit Table Knowledge with \textit{Question-Then-Pinpoint} Reasoner~for Insightful Table Summarization}

\author{
    Kwangwook Seo~~~
    Jinyoung Yeo~~~
    Dongha Lee\thanks{\; Corresponding author}\\
    Yonsei University\\
    \texttt{\{tommy2130,jinyeo,donalee\}@yonsei.ac.kr}\\   
}

\begin{document}
\maketitle
\begin{abstract}
Implicit knowledge hidden within the explicit table cells, such as data insights, is the key to generating a high-quality table summary.
However, unveiling such implicit knowledge is a non-trivial task. 
Due to the complex nature of structured tables, it is challenging even for large language models (LLMs) to mine the implicit knowledge in an insightful and faithful manner. 
To address this challenge, we propose a novel table reasoning framework \textit{\textbf{Question-then-Pinpoint}}. 
Our work focuses on building a plug-and-play table reasoner that can self-question the insightful knowledge and answer it by faithfully pinpointing evidence on the table to provide explainable guidance for the summarizer. 
To train a reliable reasoner, we collect table knowledge by guiding a teacher LLM to follow the coarse-to-fine reasoning paths and refine it through two quality enhancement strategies to selectively distill the high-quality knowledge to the reasoner.
Extensive experiments on two table summarization datasets, including our newly proposed \textbf{\textsc{InsTaSumm}}, validate the general effectiveness of our framework.\footnote{Our code and dataset are available at \url{https://github.com/tommyEzreal/QtP}.}

\end{abstract}

\section{Introduction}
\label{sec:intro}

Table data has emerged as pivotal repositories of knowledge in facilitating data analysis, offering concise and structured representation of information.
As comprehending the complex table can be time-consuming for human, text generation systems that can accurately summarize a provided table have the potential to greatly enhance  the process of obtaining data insights. 

In the task of table summarization~\citep{lebret-etal-2016-neural,suadaa-etal-2021-towards, moosavi2021scigen}, a straightforward solution is to use neural model as an end-to-end summary generator.
However, the model struggles to capture all necessary information in an end-to-end approach.
The problem lies in that unlike table question answering tasks~\citep{pasupat-liang-2015-compositional, zhong2018seqsql} where explicit guidance (\textit{i.e.}, input query) to search the answer is given, the table summarization task lacks direct control on what aspect of information should be searched from the table.
Therefore, it is challenging for the model to decide a favorable choice of implicit evidence required for summarization only from the table input. 
 
\begin{figure}[!t]
    \centering
    \includegraphics[width=1\columnwidth]{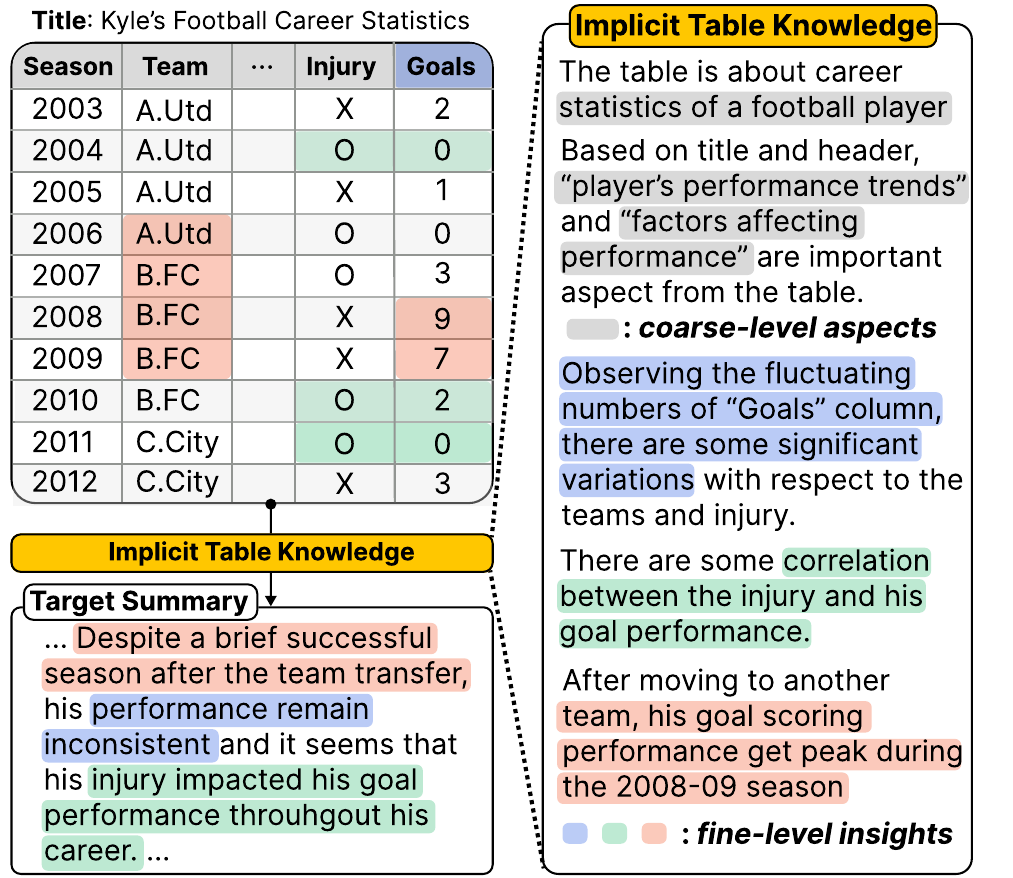}
    \caption{An example of implicit table knowledge that should be unveiled from explicitly stated table cells to generate the target summary.
    }
    \label{fig:motivating} 
\end{figure} 
A line of research address this \textit{uncontrollability} problem by injecting knowledge as a mediator to guide what kind of table contents should be stated in the text description. 
Most works adopt symbolic operations as guidance for sampling knowledge from the table to enhance the logical reasoning ability of the model. 
These symbolic operations include executable programs which mimics SQL~\citep{liu-etal-2022-plog, liu2022tapex, zhao-etal-2023-loft} or Logical Types~\citep{zhao-etal-2023-investigating, perlitz2023diversity} which categorize information seeking queries in several predefined types, serving as a control for knowledge collection.
Some works~\citep{su-etal-2021-shot-table,guo2024adapting,guo2024towards} use table as query for retrieving relevant knowledge from external source (\textit{e.g.}, KB) to supplement the insufficient table knowledge of the language model.
However, these approaches suffer from limited knowledge in terms of both diversity and coverage. 
Since symbolic operations heavily rely on rule-based sampling or predefined types, they remain insufficient for delivering comprehensive information within individual tables. 
Additionally, relying on external knowledge sources requires a rigorous assumption about the completeness of the knowledge base, which can lead to limited coverage of the required knowledge. 

In this paper, we aim to unveil the implicit table knowledge by using the reasoning ability of the large language models (LLMs). 
As shown in Figure~\ref{fig:motivating}, the implicit knowledge required for generating the target summary should encompass multiple aspects of information that are scattered across table cells.
We argue that by facilitating LLMs to directly mining these knowledge from the table with multiple reasoning paths, we can represent more diverse knowledge that can be used as informative insights to support the table summarization. 

Despite the potential effectivness of this approach, there are two major challenges in using LLM as a knowledge miner. 
\textbf{(1) Low insightfulness}: As for the complex nature of structured table, the in-depth knowledge related to the target summary are not explicitly stated in the table cells. However LLMs tend to focus more on explicit textual cues~\citep{chae2023dialogue} which often leads to surface-level realization and making it challenging to capture the insightful knowledge. 
\textbf{(2) Low faithfulness}: Since information that needs to be selected from the table to derive the reliable knowldege are scattered and hidden among irrelevant information, these often act as distracting factor or noise~\citep{patnaik2024cabinet}, leading to hallucination. 

To address the aformentioned challenges, we introduce a novel table reasoning framework, \textit{insightfully \textbf{Question} \textbf{then} faithfully \textbf{Pinpoint}}.
The key idea of our framework is to build a plug-and-play table reasoner that can self-quesiton-and-answer the insightful knowledge by faithfully pinpointing the evidence on the table to guide the table summarizer.
To train the reasoner, we collect training data from teacher LLM which follows coarse-to-fine reasoning paths to mine in-depth knowledge required for summarization. 
During this step, we refine the data using two quality enhancement strategies to selectively distill the high-quality knowledge that is helpful to train a reliable reasoner. 
                 
For effective demonstration of our framework, we conduct experiments on two table summarization datasets, including our newly proposed dataset \textbf{\textsc{InsTaSumm}}.
Since our table reasoner can be pluged-and-played to different variants of table summaization models, we validate that our framework can be applied to both fine-tuned and zero-shot summarization models with significant improvement.
In addition, we evaluate our framework under out-of-domain setting, showing its robustness in diverse real-world scenarios.

\section{Preliminaries}
\label{sec:prelim}
\paragraph{Input Table Serialization}
Following the recent works that employ language model on table-related tasks~\citep{chen-2023-large,zhao2023qtsumm, zhao-etal-2023-investigating}, we serialize the table input into a flattened sequence.
Specifically, we use a vertical bar (|) to separate headers and cells in different columns, and a newline along with the row index to separate rows.
This approach enables the direct input of structured tables into the language model.
For example, the input table $t$ is serialized as follows:

\textit{col : <header 1> | <header 2> | … row 1 : <cell value> | <cell value> | … row 2: <cell value> | <cell value> | … }




\begin{figure*}[!ht]
    \centering
    \includegraphics[width=1\textwidth]{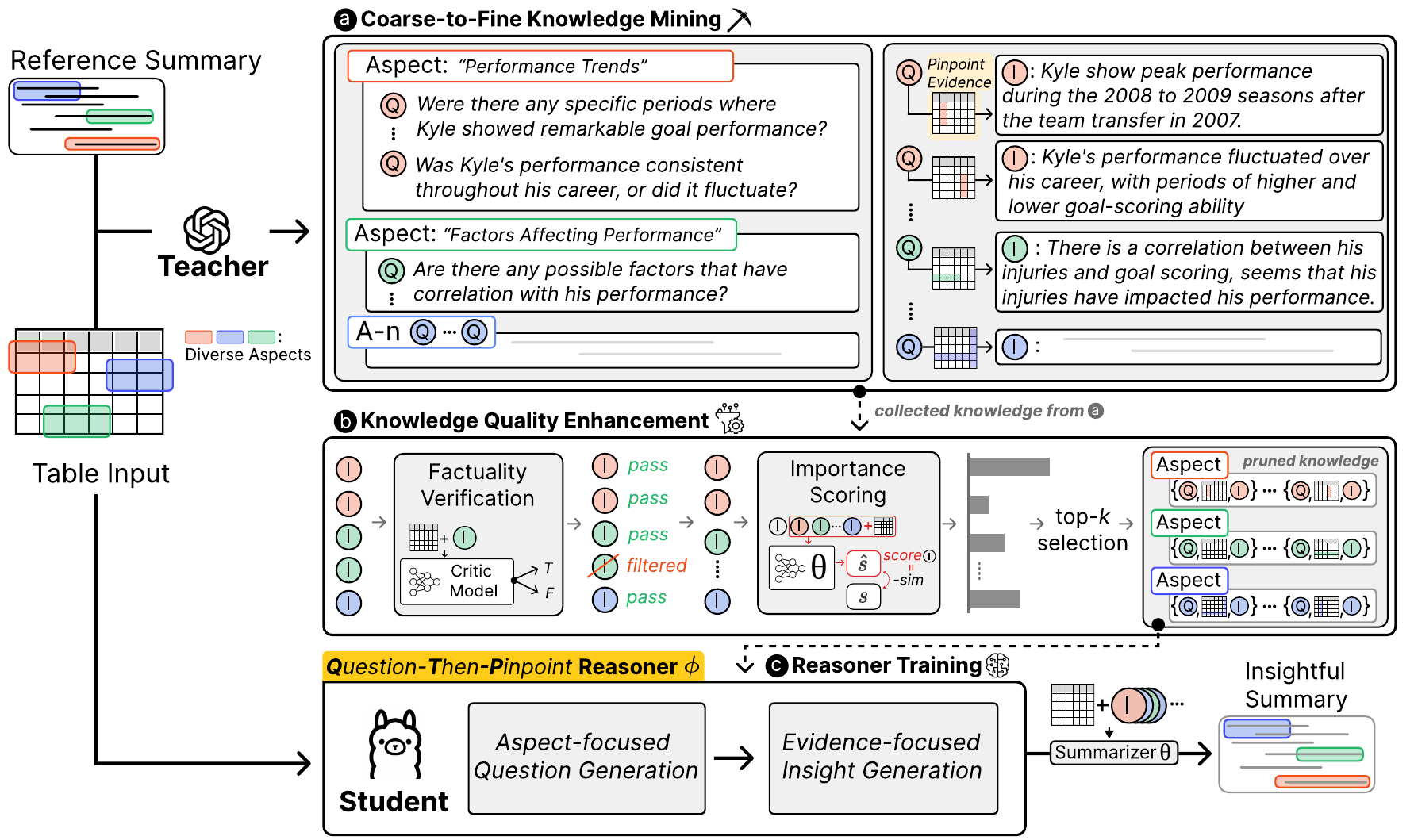}
    \caption{Overview of our framework. We (a) leverage LLM to collect diverse aspects of  knowledge from the table and reference summary. For the collected knowledge, we (b) apply two quality enhancement strategy to construct a high-quality dataset for (c) training a reliable table reasoner $\phi$. 
    During the inference phase (bottom right), the output insight $\mathcal{I}$ from the reasoner is provided to the summarizer $\theta$ as an additional input to guide the summarization. 
    }
    \label{fig:overview} ~
\end{figure*} 
\paragraph{Problem Formulation}
Existing studies on table-to-text generation~\citep{liu-etal-2022-plog, zhao-etal-2022-reastap} have mainly focused on pretraining the model with synthetic reasoning examples, and further fine-tuning them on downstream tasks in an end-to-end manner. 
Despite their progress, thsese models often struggle with generalizing to unseen domains~\citep{chen-2023-large} and consider intermediate knowledge only as a latent factor in the generation process, which poses issues of explainability. 

Inspired by recent efforts~\citep{ye2023large,Cao2023APIAssistedCG,kim-etal-2024-verifiner, ko2024evidencefocused} that incorporate auxiliary reasoning agents to empower the model's ability for downstream tasks, we extend the end-to-end table summarization setting by incorporating a table reasoner as an additional component in the generation process. This addition enables the externalization of implicit knowledge, which serve as explainable guidance for table summarization. 
Formally, given the serialized input table $t$, our reasoner focuses on generating table insights $\mathcal{I}$ as additional input knowledge to help summarizer $\theta$ in predicting the summary $s$:
\begin{equation}
    s  \sim P_{\theta}( \cdot | t, \mathcal{I})
\end{equation}

\section{Proposed Framework}
\label{sec:method}

In this section, we propose a novel table reasoning framework \textit{Question-Then-Pinpoint}, which focuses on building a
table reasoner that can provide faithful insights supportive for summarization.
The overall framework is illustrated in Figure~\ref{fig:overview}. 

\subsection{Coarse-to-Fine Knowledge Mining}
The goal of this step is to augment existing end-to-end table summarization training corpora $D = \{(t,s)\}$ with implicit table knowledge.
To mine the knowledge that is helpful to infer the target summary, we leverage the capability of LLM to rationalize the target summary from the given table.
As simply prompting LLM to generate intermediate reasoning paths often leads to surface-level knowledge, we provide several steps as checkpoints that guide the coarse-to-fine reasoning path for generating in-depth knowledge from the table.

Specifically, given the table $t$ and reference summary $s$, the teacher model $LLM_{teacher}$ first extracts coarse-level aspects $\mathcal{A} = \{a_n\}_{n=1}^{N}$, where $a_n$ represents one of the abstract topics across diverse aspects in the table.
Then, conditioned on $\mathcal{A}$, $\mathcal{Q}=\{\mathcal {Q}_{n}\}_{n=1}^{N}$ is generated, where $\mathcal{Q}_n$ is a set of fine-level questions for each $a_n$ to query the information that should be captured from $t$. 
\begin{equation}
    \mathcal{A}, \mathcal{Q} =LLM_{teacher}(t, s)
\end{equation}

After generating fine-level questions, we generate corresponding
insights as answers for each question, along with relevant cell
evidence to search for insights from the table.
Specifically, we prompt LLM to answer the given questions $\mathcal{Q}$ to generate corresponding insights $\mathcal{I} = \{\mathcal{I}_{n}\}_{n=1}^{N} $. These insights are obtained by pinpointing the cell evidence $\mathcal{E} = \{\mathcal{E}_n\}_{n=1}^{N}$, where $\mathcal{E}$ is a set of relevant cell information that provides explicit evidence from  $t$ to answer $\mathcal{Q}$.
By focusing on explicit evidence, the model can faithfully capture implicit insights while avoiding distractions from irrelevant information.
\begin{equation}
    \mathcal{E}, \mathcal{I} = LLM_{teacher}(t, s,\mathcal{Q})
\end{equation}

\subsection{Knowledge Quality Enhancement}
Symbolic knowledge distillation requires a strong teacher model to maximize the quality of the generated knowledge~\citep{west-etal-2022-symbolic}.
However, some knowledge generated by LLMs may not align with the source table, resulting in unfaithful or redundant information that may harm the helpfulness of knowledge in summary generation. 
Therefore, we propose two knowledge quality enhancement strategies to filter the low-quality knowledge generated by the teacher model and selectively distill the high-quality knowledge to the student reasoner.

\paragraph{Factuality Verification}
Despite the clear pinpointing of evidence, some of the generated knowledge might still be unfaithful and not aligned with the source table.
To effectively filter out this unfaithful knowledge, we adopt a critic model $\textbf{C}$ to classify the counterfactual knowledge.
Specifically, we use TAPEX~\citep{liu2022tapex} trained on TabFACT~\citep{Chen2020TabFact:} dataset as the critic model to verify the generated insights. Given the source table $t$ and insight set $\mathcal{I}$, the model $\textbf{C}$ do the binary classification on $i(\in \mathcal{I})$ to annotate each as consistent from $t$ or not.
We then filter out those counterfactual insights from the training dataset.

\paragraph{Importance Scoring}
In the task of table summarization, as the provided input data should be mapped into the target in an encapsulated form, simply providing additional inputs may harm the summary output. 
That is, even if the additional knowledge is faithful, some of them might not be helpful to generate the target summary. 
\begin{algorithm}[h]
\small
\caption{Importance Scoring}
\label{alg:insight_importance_scoring}
\textbf{Input} Original insight set $\mathcal{I}$,  Similarity measure $sim$, Table summarizer $\theta$, Source table $t$, Reference summary $s$ 
\begin{algorithmic}[1]
\Function{ImportanceScoring}{$\mathcal{I}$, $sim$, $\theta$, $t$, $s$}

    \State $scores \gets \{\}$
    \For{each $i \in \mathcal{I}$}
        \State $\mathcal{\Tilde{I}} \gets \mathcal{I} \setminus \{i\}$
        \State $\hat{s} \gets \theta(\mathcal{\Tilde{I}}, t)$
        \State $score \gets - sim(\hat{s}, s)$
        \State $scores[i] \gets score$
    \EndFor \Comment{Loop ends after processing all $i$ in $\mathcal{I}$}
    \State \textbf{return} $scores$
\EndFunction
\end{algorithmic}
\end{algorithm}






To address this challenge, we examine the helpfulness of generated knowledge by scoring the importance of each insight (Algorithm~\ref{alg:insight_importance_scoring}). 
We assume that by iteratively evaluating the impact of removing each insight when inferring the target summary, we can check whether each insight is actually helpful in leading the model to output the target summary. 
Specifically, we first make subset $\mathcal{\Tilde{I}}$ by removing $i(\in \mathcal{I})$ from the original set and infer table summary $\hat{s}$ from the source table $t$ conditioned on $\mathcal{\Tilde{I}}$. 
We then measure the semantic similarity between the generated summary and the reference summary to compute the importance score for each ablated $i$ with respect to the negative of the similarity. 

After repeating above process until all insights are scored, the top-$k$ insights are selected from each $\mathcal{I}_n (\subset \mathcal{I})$ for constructing the pruned training set. 
In the end, we construct an augmented  training set $D' = \{(t, s, (\mathcal{A},\mathcal{Q}, \mathcal{E}, \mathcal{I}))\}$ by mining and pruning the table knowledge $(\mathcal{A},\mathcal{Q},\mathcal{E},\mathcal{I})$.
We provide an example of processed dataset in Table~\ref{tab:knowledge_example}. 



{\renewcommand{\arraystretch}{1.1}
    \begin{table}[t] \begin{center}
    \small
    \setlength{\tabcolsep}{3pt}
    \begin{tabularx}{\linewidth}{X}
        \small
        \\
        \toprule

         \textbf{Input Table $t$:}  List of The Real Housewives of New Jersey episodes - Season 9 (2018–19) 
        
        [col] : No. overall | No. in season | Title | Original air date | U.S. viewers (millions)
        [row 1] ... \\
        \midrule
        \xspace\xspace \textbf{Table Knowledge}: $(q_1, e_1, i_1), (q_2, e_2, i_2), ... \in (\mathcal{Q}, \mathcal{E}, \mathcal{I})$

        $\bullet$ \textbf{Aspects} $\mathcal{A}$: $a_1$:Episode Highlights, $a_2$:Viewership Trends\\
        $\bullet$ \textbf{Questions} $\mathcal{Q}$: \\ $q_1$:What are some of the standout moments or highlights from the episodes with the highest viewership? \\$q_2$:Are there any noticeable patterns or fluctuations in viewership numbers across different episodes? ... \\
        $\bullet$ \textbf{Evidences} $\mathcal{E}$: \\ $e_1$:col(Title, U.S.viewers), row(13)\\ $e_2$:col(No.in season, U.S.viewers), row(3,8,13) ... \\
        $\bullet$ \textbf{Insights} $\mathcal{I}$: \\ $i_1$:The \textcolor{magenta}{standout moment is Episode 13}, titled "Camels, Cabo \& Catfights" \textcolor{magenta}{attracted highest viewership} of 1.40 million.\\ $i_2$:There are \textcolor{teal}{fluctuations in viewership numbers across different episodes}, with \textcolor{teal}{upward trends} from certain episode. ... \\        
        \midrule
         \xspace\xspace \textbf{Reference Sumamry} $s$: 
        ... \textcolor{teal}{Viewership numbers fluctuate} throughout the season with some variations. Notably, there is a noticeable \textcolor{teal}{upward trend} in viewership \textcolor{magenta}{towards later episodes, culminating in Episode 13}, titled "Camels, Cabo \& Catfights," which attracted the \textcolor{magenta}{highest viewership of 1.40 million}. ... These patterns reflect audience engagement and preferences throughout the season, indicating particular episodes that resonated more strongly with viewers.\\
        \bottomrule
    \end{tabularx}
    \vspace{-3pt}
    \caption{
        An example of training set $D'$.
    }
    \vspace{-12pt}
    \label{tab:knowledge_example}
\end{center}\end{table}}
\subsection{ \textsc{QtP} Reasoner Training}
Using the annoated dataset $D'$, we train \textit{\textbf{Q}uestion-\textbf{t}hen-\textbf{P}inpoint} \textbf{Reasoner} with two different objectives.
We employ a single student model\footnote{In this work, we choose Llama2-7b~\citep{touvron2023llama} as the backbone model for reasoner} for both question generation and insight generation, which are jointly trained on two instruction-tuning tasks.

\paragraph{Aspect-focused Question Generation Task}
The question generation task aims to generate fine-level questions to seek implicit knowledge from the table. Formally, given the source table $t$, our reasoner model $\phi$ is optimized to generate the sequence of $(\mathcal{A}, \mathcal{Q})$ pair by using the causal language modeling objective: 
\begin{equation}
\small
    \mathcal{L}_{\text{QG}}(t, \mathcal{A}, \mathcal{Q}) = - \log p_\phi(\mathcal{A,Q}| t)
\end{equation}


\paragraph{Evidence-focused Insight Generation Task}
The insight generation task aims to predict insights by answering the given question, focusing on cell evidence.
With the sequential prediction of both cell evidence $e$ and the corresponding $i$, the model can learn to capture faithful insights based on the pinpointed evidence.
Given a question $q$ and source table $t$, the insight generation module aims to generate insight $i$ by answering the given $q$: 
\begin{equation}
\small
\mathcal{L}_{\text{IG}}(t, \mathcal{Q}, \mathcal{E}, \mathcal{I}) = - \underset{(q, e, i)\in (\mathcal{Q},  \mathcal{E}, \mathcal{I})}{\sum}\log p_\phi(e,i| t, q)
\end{equation}



The final objective function of \textsc{QtP} reasoner is the combination of question and insight generation:
\begin{equation}
\small
\begin{split}
    &\mathcal{L}_{\text{Reasoner}} = \\
    &\underset{(t, s, (\mathcal{A}, \mathcal{Q}, \mathcal{E}, \mathcal{I}))\sim D'}{\mathbb{E}} \left[\mathcal{L}_{\text{QG}}(t, \mathcal{A}, \mathcal{Q}) + \mathcal{L}_{\text{IG}}(t, \mathcal{Q}, \mathcal{E}, \mathcal{I})\right]
\end{split}
\raisetag{39pt}
\end{equation}

\section{Experiments}
\label{sec:exp}
We conduct extensive experiments on summarization to demonstrate how the insights from \textsc{QtP} Reasoner provide useful guidance to the summarizer in generating high-quality table summaries.
\subsection{Datasets}
\paragraph{In-domain} We first evaluate the performance on the test set held-out from the dataset for training \textsc{QtP} Reasoner. 
Since existing open-domain table-to-text generation datasets mainly focus on sentence-level generation~\citep{chen-etal-2020-logical,parikh-etal-2020-totto} or are limited to specific domain~\citep{liang-etal-2009-learning, wiseman-etal-2017-challenges, suadaa-etal-2021-towards}, we require a more comprehensive testbed for evaluating our framework. 
Therefore, we build a refined version of an existing dataset named \textbf{\textsc{InsTaSumm}}, which focuses on generating \textbf{Ins}ightful \textbf{Ta}ble \textbf{Summ}ary solely from the input table in a paragraph format.

We adopt QTSumm~\citep{zhao2023qtsumm} as a source dataset to construct \textsc{InsTaSumm}. QTSumm is a query-focused table summarization dataset, collected by human-annotated multiple queries and summaries for a single table input. 
As QTSumm considers informativeness when curating queries and covers diverse aspects with multiple query-summary pairs for each table, it consists of rich and in-depth information in the annotated descriptions compared to general table-to-text datasets. 
Hence, we construct \textsc{InsTaSumm} to comprise a paragraph-form summary for each individual table by aggregating diverse query-focused summaries from QTSumm.  
Instead of simply concatenating, we aggregate them into a single summary by prompting GPT-4 to verbalize it in a more fluent form.
We provide detailed statistics of \textsc{InsTaSumm} in Appendix~\ref{sec:dataset_detail}.

\paragraph{Out-of-domain} To further evaluate the generalizability of our framework, we choose \textbf{SciGEN}~\citep{moosavi2021scigen} as an out-of-domain dataset. 
SciGEN is a domain-specific table-to-text dataset, which is collected from scientific articles. 
It requires intensive reasoning to generate the long-form description from the given table.
We use the test split of the medium setting for the experiment.

\subsection{Evaluation Metrics}
To evaluate the table summarization performance from multiple perspectives, we employ various automated evaluation metrics at four different levels. 
\textbf{(1) Surface-level}: We adopt SacreBLEU, ROUGE, METEOR, BERTScore, and A3CU~\citep{zhao2023qtsumm,post-2018-call,liu-etal-2023-towards-interpretable} to evaluate both lexical overlap and contextual similarity between the reference and inferred summary. 
\textbf{(2) Faithfulness-level}: Following the previous works~\citep{liu-etal-2022-plog,zhao2023qtsumm,zhao-etal-2023-investigating}, we use TAPAS-Acc and GPT4-Acc to evaluate the factual correctness of the generated summary. 
\textbf{(3) Insightfulness-level}: We use G-EVAL~\citep{liu2023geval} approach to evaluate the analytical depth of each summary. Specifically, we prompt GPT-4 to evaluate the insightfulness of the generated summary for the given table and summary pair in 1 to 5 Likert scale and report the average score. 
\textbf{(4) Pairwise quality comparison}: Following ~\citet{dubois2024lengthcontrolled}, we conduct a pairwise comparison where we present a source table and two summaries made by different models and ask GPT-4 to choose one based on diverse criteria. We adopt three criteria: which table summary is more \textit{natural}, \textit{comprehensive}, and  \textit{informative}.  
We provide the details of each metric and all the prompts used in the evaluation in Appendix~\ref{sec:evaluation_detail}.
\begin{table*}[t!]
\centering
\resizebox{\linewidth}{!}{
\begin{tabular}{clcccccccc}
\\
\toprule
\multicolumn{2}{l}{
\textbf{\textsc{InsTaSumm}} (\textit{In-domain})} \\

\midrule
  \multirow{2.5}{*}{\textbf{Type}} &
  \multirow{2.5}{*}{\textbf{Methods}} &
  \multicolumn{5}{c}{\textbf{Surface-level}} &
  \multicolumn{2}{c}{\textbf{Faithfulness-level}} &
  \multicolumn{1}{c}{\textbf{Insightfulness-level}} \\
  \cmidrule(lr){3-7} \cmidrule(lr){8-9} \cmidrule(lr){10-10}

& & S-BLEU & ROUGE-L & METEOR & BERTScore & A3CU & TAPAS-Acc & GPT4-Acc & G-EVAL  \\
\midrule
\multirow{6}{*}{\shortstack[c]{\\\emph{\textbf{fine-tuned}} \\\emph{summarizer} }}
 & ReasTAP~\citep{zhao-etal-2022-reastap} & 9.34 & \textbf{33.70} & 33.45 & 87.67 & 29.61 & 68.18 & 69.55 & 2.79\\
& \hspace{0.1cm}+ CoT Reasoner & 9.34 & 32.24 &  33.88 & 87.74 & 27.36 & 69.31 & 65.49 & 2.87  \\
& \hspace{0.1cm}+ Plan-and-Solve Reasoner & 9.03 & 31.91 & 34.62  & 87.60 & 30.51 & 67.04 & 70.82 & 2.43  \\
& \hspace{0.1cm}+ Logical Type Reasoner & 9.48 & 32.70 & 33.99  & 87.85 & 27.70 &  70.45 & 72.43 & 2.50 \\
& \hspace{0.1cm}+ SQL Reasoner& 9.11 & 32.38 & 33.30 & 87.67 & 27.35 & 70.90  & \textbf{73.72} & 2.74 \\
&\cellcolor{gray!20} \hspace{0.1cm}+ \textbf{\textsc{QtP} Reasoner (ours)}  & \cellcolor{gray!20}\textbf{10.83} &\cellcolor{gray!20} 32.68 & \cellcolor{gray!20}\textbf{36.35} & \cellcolor{gray!20}\textbf{88.25}  & \cellcolor{gray!20}\textbf{31.25}  & \cellcolor{gray!20}\textbf{75.68} &\cellcolor{gray!20} 73.04 & \cellcolor{gray!20}\textbf{3.05} \\

\midrule
\multirow{6}{*}{\shortstack[c]{\\\emph{\textbf{fine-tuned}} \\\emph{summarizer} }}
& Llama-2-7b~\citep{touvron2023llama} & 16.47 & 29.70 & 36.05 & 89.23 & 26.72 & 78.86 & 74.61 & 2.95 \\
& \hspace{0.1cm}+ CoT Reasoner & 16.43 & 28.92 & 34.20  & 88.93  & 27.93 & 77.27 & 78.65 & 3.03  \\
& \hspace{0.1cm}+ Plan-and-Solve Reasoner& 17.39 & 29.98 & 35.18  & 88.78  & 26.49 & 78.40 & 77.89 & 2.87  \\
& \hspace{0.1cm}+ Logical Type Reasoner & 16.02 & 28.28 & 36.01  & 88.54 & 23.72 & 81.81 & 76.29 & 2.80 \\
& \hspace{0.1cm}+ SQL Reasoner& 17.70 & 28.35 & 33.24 & 88.62 & 24.57 & 79.54 & 80.48 & 2.79 \\
& \cellcolor{gray!20}\hspace{0.1cm}+ \textbf{\textsc{QtP} Reasoner (ours)} & \cellcolor{gray!20}\textbf{19.48} & \cellcolor{gray!20}\textbf{31.79} & \cellcolor{gray!20}\textbf{40.29}  & \cellcolor{gray!20}\textbf{89.76}  & \cellcolor{gray!20}\textbf{30.28} & \cellcolor{gray!20}\textbf{85.90} & \cellcolor{gray!20}\textbf{84.83} & \cellcolor{gray!20}\textbf{3.34} \\

\midrule
\multirow{6}{*}{\shortstack[c]{\\\emph{\textbf{zero-shot}} \\\emph{summarizer} }}
& Mistral-7b~\citep{jiang2023mistral} & 7.31 & 21.41 & 29.42  & 85.91 & 18.25 & 68.40 & 64.46 & 1.98 \\
& \hspace{0.1cm}+ CoT Reasoner & 7.60 & 20.26 & 26.26 & 86.46 & 15.41 & 70.90 & 73.24 & 2.24 \\
& \hspace{0.1cm}+ Plan-and-Solve Reasoner& 8.02 & \textbf{22.63} & 29.94  & 86.61 & 16.89 & 70.54 & 69.31 & 2.85 \\
& \hspace{0.1cm}+ Logical Type Reasoner & 7.33 & 20.53 & 28.03  & 86.41 & 16.53 & 71.36 & 73.63 & 2.75 \\
& \hspace{0.1cm}+ SQL Reasoner& 7.85 & 20.52 & 27.19 & 86.49 & 17.36 & 72.04 & 74.80 & 2.36 \\
& \cellcolor{gray!20}\hspace{0.1cm}+ \textbf{\textsc{QtP} Reasoner (ours)} & \cellcolor{gray!20}\textbf{8.53} &\cellcolor{gray!20} 21.49 & \cellcolor{gray!20}\textbf{34.95}  & \cellcolor{gray!20}\textbf{87.44}  & \cellcolor{gray!20}\textbf{21.02} & \cellcolor{gray!20}\textbf{82.72} & \cellcolor{gray!20}\textbf{78.04} & \cellcolor{gray!20}\textbf{3.18} \\

\midrule
\multirow{8}{*}{\shortstack[c]{\\\emph{\textbf{zero-shot}} \\\emph{summarizer} }}
& GPT-3.5-turbo~\citep{openai2023chatgpt} & 7.64 & 23.45 & 28.43  & 87.12 & \textbf{23.96} & 72.49 & 76.74 & 1.94 \\

& \hspace{0.1cm}+ CoT Reasoner& 8.69 & 23.90 & 27.06  & 87.62 & 20.90 & 65.90 & 74.40 & 2.10 \\
& \hspace{0.1cm}+ Plan-and-Solve Reasoner& 9.77 & 24.10 & 29.04 & 87.77 & 21.87 & 65.45 & 71.11 & 2.74\\
& \hspace{0.1cm}+ Logical Type Reasoner& 8.28 & 23.38 & 26.39  & 87.39 & 19.53 & 67.95 & 71.66 & 2.04 \\
& \hspace{0.1cm}+ SQL Reasoner& 8.92 & 24.22 & 26.69 & 87.61 & 22.50 & 70.68 & 79.35 & 2.32 \\

& \cellcolor{gray!20}\hspace{0.1cm}+ \textbf{\textsc{QtP} Reasoner (ours)} & \cellcolor{gray!20}\textbf{11.38} & \cellcolor{gray!20}\textbf{25.04} & \cellcolor{gray!20}\textbf{34.17} & \cellcolor{gray!20}\textbf{88.07} & \cellcolor{gray!20}23.48 & \cellcolor{gray!20}\textbf{88.63} & \cellcolor{gray!20}\textbf{85.33} & \cellcolor{gray!20}\textbf{3.52} \\
\cmidrule(lr){2-10}
& \hspace{0.1cm}+ Self-Gen Knowledge & 9.31 & 24.08 & 29.35 & 87.53 & 21.23 & 86.59 & 84.26 & 3.40 \\

& \hspace{0.1cm}+ \textsc{ORACLE} Knowledge & 13.56 & 26.46 & 36.14 & 88.63 & 26.48 & 89.09 & 87.41 & 3.34 \\
\bottomrule
\end{tabular}
}
\caption{In-domain summarization results on \textsc{InsTaSumm} testset.}
\label{tab:main_table}
\end{table*}
\subsection{Table Summarizer}
We consider both fine-tuned and zero-shot table summarization models for assessing the helpfulness of our reasoner in diverse scenarios.  
\textbf{(1) Fine-tuned Summarizer:} We consider two foundation models, \textbf{ReasTAP}~\citep{zhao-etal-2022-reastap} and \textbf{Llama-2-7b-chat}~\citep{touvron2023llama}. 
\textbf{(2) Zero-shot Summarizer:} For zero-shot evaluation, we consider two large-scale models, \textbf{GPT-3.5-turbo}~\citep{openai2023chatgpt}, \textbf{Mistral-7b}~\citep{jiang2023mistral}.
We provide details on each model in Appendix~\ref{sec:Table_summarizer}.

Specifically, for both scenarios, we provide the knowledge generated by \textsc{QtP} Reasoner as an additional input. 
For fine-tuned summarizers, we augment the input during both the training and inference phases, while for zero-shot summarizers, we provide the knowledge during the inference.

\begin{figure}[!t]
    \centering
    \includegraphics[width=1\columnwidth]{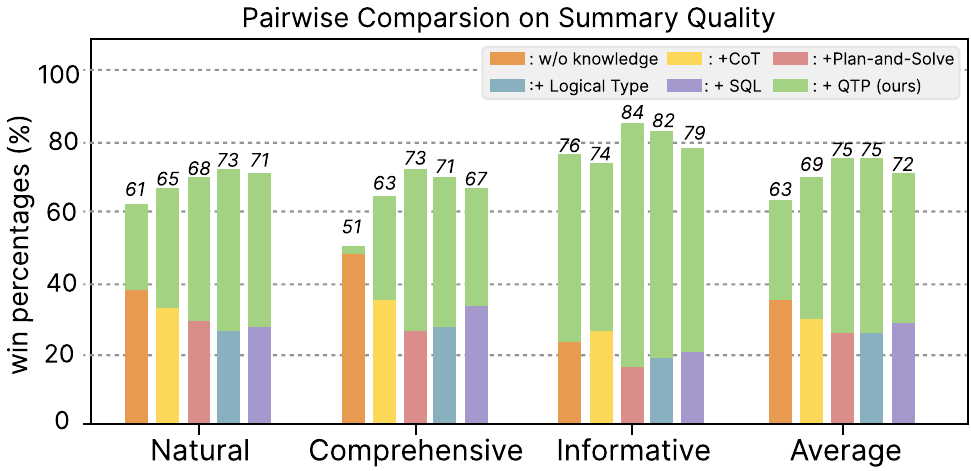}
    \caption{Pairwise \textbf{summary quality} comparison results on \textsc{InsTaSumm} using GPT-3.5 as backbone summarizer. We report the win percentage of \textsc{QtP} Reasoner.
    }
    \label{fig:pairwise_summary} 
\end{figure} 
\subsection{Baselines}
To evaluate how knowledge affects the performance of summarization, we compare \textsc{QtP} with the following baselines.
\textbf{(1) Without knowledge:} We first consider the end-to-end baseline, where the summarizer directly predicts the target summary without externalization of implicit knowledge. \textbf{(2) Generate knowledge with step-by-step reasoning:} We consider another baseline that uses generic LLM reasoning to generate knowledge for augmenting the summarizer. 
Specifically, we implement two knowledge models, including CoT Reasoner~\citep{yang-etal-2024-effective-distillation, NEURIPS2022_zeroshotcot} and Plan-and-Solve (P\&S) Reasoner~\citep{wang2023planandsolve}, that generate implicit knowledge based on step-by-step reasoning.
\textbf{(3) Generate knowledge with symbolic reasoning:} 
We then consider task-specific reasoners as baselines that guide knowledge generation with logical table operations.
For Logical Type (LT) Reasoner, we adopt 9 predefined operation types\footnote{Following ~\citet{zhao-etal-2023-investigating}, we adopt Aggregation, Negation, Superlative, Count, Comparative, Ordinal, Unique, All, and Surface} as control for knowledge generation, following \citet{zhao-etal-2023-investigating,perlitz2023diversity}.
For SQL Reasoner, we use SQL queries as guidance for generation, following the concept of \citet{liu-etal-2022-plog,liu2022tapex,ye2023large,cheng2022binding}.

\begin{table*}[t!]
\centering
\resizebox{\linewidth}{!}{
\begin{tabular}{clcccccccc}
\toprule
\multicolumn{2}{l}{\textbf{\textsc{SciGEN}} (\textit{Out-of-domain})} \\

\midrule
\multirow{2.5}{*}{\textbf{Type}} &
\multirow{2.5}{*}{\textbf{Methods}} &
\multicolumn{5}{c}{\textbf{Surface-level}} &
\multicolumn{2}{c}{\textbf{Faithfulness-level}} &
\multicolumn{1}{c}{\textbf{Insightfulness-level}} \\
\cmidrule(lr){3-7} \cmidrule(lr){8-9} \cmidrule(lr){10-10}

& & {S-BLEU} & {ROUGE-L} & METEOR & BERTScore & {A3CU} & {TAPAS-Acc} & { GPT4-Acc} & G-EVAL  \\
\midrule
\multirow{6}{*}{\shortstack[c]{\\\emph{\textbf{zero-shot}} \\\emph{summarizer} }}
& Mistral-7b~\citep{jiang2023mistral} & 1.89 & 14.09 & 20.09 & 82.42 & 6.91 & 71.19 & 69.80 & 1.73  \\
& \hspace{0.1cm}+ CoT Reasoner & 1.87 & 13.90 & 19.49  & 83.24 & 6.06 & 68.88 & 67.44 & 2.56 \\
& \hspace{0.1cm}+ Plan-and-Solve Reasoner& 2.08 & \textbf{14.13} & 22.03  & 83.33 & 8.03 & 69.36 & 69.53 & 3.08 \\
& \hspace{0.1cm}+ Logical Type Reasoner & 1.95 & 13.74 & 19.72 & 83.12 & 6.20 & 70.32 & 71.36 & 2.10 \\
& \hspace{0.1cm}+ SQL Reasoner& 1.90 & 13.21 & 19.50 & 83.07 & 7.49 & 72.73 & 74.85 & 2.03 \\
& \cellcolor{gray!20} \hspace{0.1cm}+ \textbf{\textsc{QtP} Reasoner (ours)}  & \cellcolor{gray!20}\textbf{2.22} & \cellcolor{gray!20}13.67 & \cellcolor{gray!20}\textbf{23.51}  & \cellcolor{gray!20}\textbf{83.82}  & \cellcolor{gray!20}\textbf{9.19} & \cellcolor{gray!20}\textbf{75.59} & \cellcolor{gray!20}\textbf{75.49} & \cellcolor{gray!20}\textbf{3.35} \\ 
\midrule
& GPT-3.5-turbo~\citep{openai2023chatgpt} & 2.53 & 15.53 & 19.71 & 83.64 & 7.45 & 81.40 & 76.48 & 2.05 \\
\multirow{4}{*}{\shortstack[c]{\\\emph{\textbf{zero-shot}} \\\emph{summarizer} }} 
& \hspace{0.1cm}+ CoT Reasoner& 2.41 & 15.39 & 18.38  & 84.11 & 7.07 & 75.91 & 78.44 & 3.08  \\
& \hspace{0.1cm}+ Plan-and-Solve Reasoner& \textbf{3.42} & 15.33 & 22.15  & 83.98  & 8.77 & 70.71 & 71.20 & 3.24  \\
& \hspace{0.1cm}+ Logical Type Reasoner& 2.49 & 15.20 & 18.67 & 83.89 & 6.86 & 79.96 & 81.34 & 2.54 \\
& \hspace{0.1cm}+ SQL Reasoner& 2.52 & 15.41 & 18.21 & 83.95 & 7.01 & 82.85 & 82.58 & 2.38 \\
& \cellcolor{gray!20}\hspace{0.1cm}+ \textbf{\textsc{QtP} Reasoner (ours)} & \cellcolor{gray!20}3.16 & \cellcolor{gray!20}\textbf{15.67} & \cellcolor{gray!20}\textbf{23.30} & \cellcolor{gray!20}\textbf{84.53} & \cellcolor{gray!20}\textbf{9.37} & \cellcolor{gray!20}\textbf{91.71} & \cellcolor{gray!20}\textbf{86.45} & \cellcolor{gray!20}\textbf{3.94} \\ 
\bottomrule

\end{tabular}
}
\caption{Out-of-domain summarization results on SciGEN testset.}
\label{tab:ood_table}
\end{table*}

For a fair comparison, we train all baseline reasoners with the same backbone model as \textsc{QtP} Reasoner, by distilling the reasoning ability of LLM. 
We provide more details about implementations in Appendix~\ref{sec:baselines_detail}.

\subsection{Main Results}

\paragraph{Comparison with end-to-end approach}
We first compare our approach with different variants of end-to-end summary generation.
Table~\ref{tab:main_table} shows that summary conditioned on knowledge from \textsc{QtP} reasoner significantly improves the performance of both fine-tuned and zero-shot summarizers. 
From Figure~\ref{fig:pairwise_summary}, we find that summaries conditioned on knowledge from \textsc{QtP} Reasoner tend to be more natural, comprehensive, and informative.
These suggest that augmenting the summarizer with the \textsc{QtP} Reasoner is beneficial for capturing related knowledge to produce a better-quality summary.
In addition, the consistent improvements on different backbone summarizers demonstrates the general effectiveness of our approach. 

\paragraph{Comparison with other reasoner baselines}
We compare \textsc{QtP} Reasoner with other knowledge-augmented baselines that generate knowledge with two different types of reasoning, \textit{i.e.,} step-by-step reasoning (CoT, Plan-and-Solve) and symbolic reasoning (Logical Type, SQL). 
From Table~\ref{tab:main_table} and Figure~\ref{fig:pairwise_summary}, we observe that incorporating baseline knowledge models into table summarization yields only marginal improvements, and in some metrics, it even underperforms compared to the end-to-end model.
Specifically, we find that while symbolic reasoning enhances the factual correctness of the summary, it falls short of enhancing insightfulness.
We also observe that although the Plan-and-Solve Reasoner achieves comparable performance on surface and insightfulness metrics to \textsc{QtP} Reasoner with multi-step reasoning, it still suffers from low faithfulness. 
In contrast, our model remains robust against this insightful-faithful trade-off by grounding the coarse-to-fine knowledge mining process in explicitly pinpointed evidence.

\paragraph{Comparison with oracle and self-generated knowledge}
To further demonstrate the efficacy of our approach, we compare the \textsc{QtP} Reasoner with two additional baselines. 
First, we augment the summarizer with \textsc{ORACLE} knowledge, which is obtained from the held-out test split in \textsc{InsTaSumm}, serving as the upper bound for summarization performance. 
Additionally, we introduce another setting where the teacher LLM directly generates knowledge to augment the summarizer without distillation to the student reasoner, which we refer to as Self-Generated Knowledge. 
From the results in Table~\ref{tab:main_table}, we find that \textsc{QtP} Reasoner-augmented summarizer shows comparable performance to the summarizer augmented with oracle knowledge, even without referencing the ground-truth summary. 
Moreover, augmenting the summarizer directly with knowledge generated by the teacher LLM does not result in better summary quality compared to \textsc{QtP} Reasoner. 
These results indicate that LLM-generated knowledge is not always helpful for summarization, highlighting the need for a selective distillation mechanism with a quality enhancement strategy to ensure the quality of the generated knowledge.
\begin{table}[t]
\setlength{\tabcolsep}{3pt}
\centering
\small
\begin{tabular}{l|XXXX}

\toprule
\textsc{QtP} vs.
 & CoT & {P\&S} & {LT} & SQL  \\
\midrule
Diverse & 59\%$^*$ & 57\%$^*$& 52\%\ \ \ \ & 69\%$^*$ \\
Insightful & 73\%$^*$  & 68\%$^*$ & 76\%$^*$  & 84\%$^*$ \\
Faithful & 61\%$^*$ & 66\%$^*$ & 58\%$^*$ & 51\%\ \ \ \ \\ 
\bottomrule
\end{tabular}
\caption{Human evaluation on \textbf{knowledge qulaity}. We report \textsc{QtP}'s win percentages. (*: p-value $< 0.05$)}
\label{tab:human_eval_knowledge}
\end{table}
\paragraph{Generalizability of \textsc{QtP} in out-of-domain scenario}
In Table~\ref{tab:ood_table}, we observe that our approach outperforms all baselines in out-of-domain scenarios, where the test domain is unseen during the training phase. This is attributed to \textsc{QtP} Reasoner's generalization ability, which stems from its flexibility of self-questioning the required knowledge from the unseen tables. 
While LLMs show remarkable generalization ability in diverse tasks, we find that they can still benefit from \textsc{QtP} Reasoner in capturing implicit knowledge that provides robust guidance for unseen domains.


\subsection{Analysis}

\paragraph{\textsc{QtP} produces better-quality knowledge} 
To assess \textsc{QtP} Reasoner's ability to generate implicit knowledge from tables, we conduct a human evaluation on knowledge quality, focusing on three dimensions: \textit{diversity}, \textit{insightfulness}, and \textit{faithfulness}.
We randomly sample 100 reasoner inferences on the test set of \textsc{InsTaSumm} and ask 3 different human judges to compare the knowledge from \textsc{QtP} paired with baselines.
We provide evaluation details in Appendix~\ref{sec:evaluation_detail}.
The results are shown in Table~\ref{tab:human_eval_knowledge}.
We can see that while baselines achieve comparable performance in the diversity of knowledge against \textsc{QtP}, they usually struggle to generate insightful and faithful knowledge. 
This suggests that \textsc{QtP} generates better-quality knowledge that provides more in-depth and accurate analysis.

\paragraph{Question guides the deeper analysis of the table, while evidence pinpoints faithful cues}
To analyze the role of question and cell evidence, we perform ablation studies on knowledge generation.
Specifically, we remove all questions from the dataset and train the model to generate insights along with cell evidence. Next, we ablate cell evidence and train the reasoner to directly predict insights for each question.
The results are shown in Table~\ref{tab:ablation}. 
We observe that without questions, the summary quality drops significantly, especially in surface-level and insightfulness-level metrics. When ablating cell evidence, the performance decreases especially in faithfulness-level metrics.  
From these results, we posit that question plays a crucial role in providing specific guidance to the model for searching in-depth knowledge from the table, while the role of evidence is narrowing down the search space with clear pinpointing to avoid distractions from irrelevant cell values.




\begin{table}[t]

\resizebox{\columnwidth}{!}{%
    \centering
    \begin{tabular}{lcccc}
    \toprule
    Training&
      ROUGE-L &
      BERTScore & 
      TAPAS-Acc &
      G-EVAL \\ 
    \midrule
    \textsc{QtP} (full) & \textbf{25.04} & 88.07 & \textbf{88.63} & \textbf{3.52}\\ 
    \midrule    
    \hspace{0.1cm}w/o question $\mathcal{Q}$ & 21.35 & 86.14 & 85.22 & 3.16 \\
    \hspace{0.1cm}w/o evidence $\mathcal{E}$ & 24.94 & \textbf{88.15} & 81.81 & 3.43\\
    \midrule
    \hspace{0.1cm}w/o Fact Verif. & 24.15 & 87.93 & 86.36 & 3.38  \\
    \hspace{0.1cm}w/o Impt. Scoring & 23.59 & 87.12 & 87.50 & 3.20 \\
    \bottomrule
    
    \end{tabular}
    }
    \caption{Ablation results on \textsc{InsTaSumm} using GPT-3.5-turbo as backbone summarizer. 
}
\label{tab:ablation}
\end{table}
\paragraph{Knowledge Quality Enhancement leads to better-quality summary} 
To investigate the effect of knowledge quality enhancement strategies, we construct two different training data by omitting each strategy and training different versions of the reasoner.
In Table~\ref{tab:ablation}, we find that factuality verification impacts more on faithfulness while importance scoring impacts on the surface and insightfulness metrics.
These results suggest that a quality enhancement strategy for selecting key knowledge that factually aligns with the table is essential for training a reliable knowledge model.

\paragraph{Case Study}
We present an example summary from \textsc{QtP}  paired with other baselines in Table~\ref{tab:case_summary_instasumm}.
The table shows that \textsc{QtP} provides a more comprehensive analysis and offers detailed information compared to the baselines. While the baselines merely list the facts from the table, \textsc{QtP}-generated summary is well-structured with a logical flow that transitions seamlessly from a general overview to specific details, making it easier to follow the narrative. We present more examples in Appendix~\ref{sec:case_study_appendix}.


\section{Related Work}
\label{sec:relwork}
\paragraph{Reasoning Over Table}
\label{subsec:t2t}
Enhancing the reasoning ability of models has been explored in diverse table-related tasks, including TableQA, TableFV, and Table-to-Text.
Existing works~\citep{liu-etal-2022-plog,zhao-etal-2022-reastap} have mainly focused on pretraining the model with auxiliary reasoning tasks with large-scale corpora.
Recently, some works~\citep{zhao-etal-2023-investigating,ye2023large,yang-etal-2024-effective-distillation} have shown the ability of LLMs in diverse table-based tasks through step-by-step reasoning. However, LLMs still suffer from unreliable predictions, leading to unfaithful or low-quality generation.
Inspired by the recent works on knowledge distillation~\citep{west-etal-2022-symbolic, hsieh2023distillingstepbystepoutperforminglarger}, we focus on building a reliable table reasoner, which selectively distills high-quality knowledge from the teacher LLM and helps the downstream task model as a plug-and-play module.

\paragraph{Table-to-Text Generation}
\label{subsec:t2t}
Recently, some works have adopted the knowledge-augmented approach in a table-to-text generation where the model is provided supplementary knowledge for the target text. Most of this knowledge is collected based on logical table operations~\citep{zhao-etal-2023-loft,zhao2023qtsumm,perlitz2023diversity}, 
such as Logical Form and Logical Types, or retrieved from the external knowledge source~\citep{guo2024adapting,guo2024towards} to supplement the insufficient domain knowledge of the language model. 
Nevertheless, these can suffer from being constrained to certain logical types or limited coverage of external knowledge sources.
Therefore, we propose \textsc{QtP} which uses LLM as a knowledge generator to represent more diverse and complex knowledge by directly mining the knowledge from the table.

\section{Conclusion}
\label{sec:conclusion}
In this paper, we propose \textit{Question-then-Pinpoint}, a novel table reasoning framework that builds a plug-and-play table reasoner providing faithful insights supportive for table summarization. 
To achieve this, we mine the implicit table knowledge via coarse-to-fine reasoning paths and train the reasoner to self-question-and-answer the required knowledge by pinpointing the explicit evidence.  
We conduct extensive experiments on two different datasets including our newly proposed \textsc{InsTaSumm}, and demonstrate the general effectiveness of our framework compared to the baselines. 

\section*{Limitations}
\label{sec:limitation}
Despite the remarkable performance of our approach, several limitations remain, suggesting areas for future improvement.
First limitation is the maximum sequence length limit occurred from input serialization. While we reduce noise in the table by explicitly pinpointing relevant evidence for each insight, our approach still requires to serialize the whole input table. 
However, real-world tables can be much longer than existing benchmark tables,  exceeding the input length capacity of the language model. 
To address this, we plan to explore methods to reconstruct larger input tables into a more concise form that encapsulates compact information.

Second, our method and dataset currently do not explicitly handle multiple tables or hierarchical tables (\textit{i.e.,}header or cells exhibits a multi-level structure) as input, which contain more complex structures than those used in our experiments. 
Considering these is important in terms of the applicability of our system in real-life scenarios where financial experts, such as analysts or investors, often refer to multiple hierarchical tables to obtain insightful conclusions~\citep{Zhao2022MultiHierttNR}. 
Therefore, extending our framework to effectively understand and process these hierarchical tables would be an important future direction.

Lastly, the reliable automated evaluation for the generated summary still remains as a challenge. Current table-related tasks often suffer from unreliable automated evaluation metrics that do not align well with human evaluations~\citep{liu-etal-2022-plog,liu-etal-2023-towards-interpretable}. 
While we employ certain automatic evaluation metrics (\textit{e.g.,} faithfulness-level, insightfulness -level) that assess the quality of summaries beyond mere surface-level matching to references, these metrics may contain inherent biases~\citep{dubois2024lengthcontrolled} that affect their ability to accurately measure the true faithfulness or insightfulness of the generated outputs.
Recent studies on the fine-grained evaluating LMs shows remarkable progress across diverse tasks~\citep{Jiang2023TIGERScoreTB,Zhu2023JudgeLMFL,Kim2023PrometheusIF} that better aligns with the human judgement with the customized evaluation criteria tailored for each tasks. 
These trends inspired us to develop a robust, table structure-aware evaluator that can assess the fine-grained quality of the generated outputs from diverse perspectives , aligning closely with human judgments in the future work. 

\section*{Ethical Consideration }
\label{sec:Ethics}
Texts generation output of LLMs may include harmful, biased, or offensive content.
However, we assert that in our research, this risk is largely minimized. 
The source tables and reference summaries in \textsc{InsTaSumm} are collected from QTSumm~\citep{zhao2023qtsumm}, which is a publicly available dataset and have annotated by humans. 
We also check the generated knowledge with manual elimination of toxic, offensive or biased uses of lanaguage. 
For human evaluation, we hire three different judges from Amazon Mechanical Turk and guarantee fair compensation for each judge. 
We pay \$0.15 for each unit task. 
The presented \textsc{InsTaSumm} dataset does not contain personal information that could lead to the identification of individuals or groups. 

\section*{Acknowledgement}
\label{sec:Ack}
This work was supported by the IITP grants funded by the Korea government (MSIT) (No. RS-2020-II201361; RS-2024-00457882, AI Research Hub Project), and the NRF grant funded by the Korea government (MSIT) (No. RS-2023-00244689).

\bibliography{anthology,custom}
\bibliographystyle{acl_natbib}

\appendix

\newpage
\section{Experimental Details}
\label{sec:appendix}

\subsection{QtP Reasoner Training Data Generation}
\label{sec:QTP_data_generation}
\paragraph{Coarse-to-Fine Knowledge Mining}
We prompt \texttt{GPT-3.5-turbo-0125} to generate the implicit knowledge for a given source table and the target summary with a 1-shot demonstration.
We empirically confirmed that increasing the number of demonstrations does not necessarily guarantee higher-quality training data. 
The reason for this is that providing too many examples can lead to the generation of responses that closely follow the reasoning paths presented in those examples, especially when dealing with tables from similar domains. 
This tendency reduces the diversity of questions needed to gather a variety of insights. 
Therefore, we selected demonstrations that are not overly specific to any particular reasoning path and focused more on instruction that closely align with our desired output format. 
The prompt used for knowledge generation is shown in Table~\ref{tab:crf_AQ} and~\ref{tab:crf_EI}. 
To collect sufficient knowledge candidates, we generate five questions for each aspect during the initial generation.

\paragraph{Knowledge Qulaity Enhancement}
After all sets of candidate knowledge are generated, we apply two quality enhancement strategies to filter out the low-quality knowledge.   
We adopt TAPEX~\citep{liu2022tapex} trained on TABFACT~\citep{Chen2020TabFact:}  to verify the incorrect insights. 
TAPEX enhances the pre-training of the BART model with a vast corpus of synthetic SQL query execution data, improving its table comprehension and reasoning abilities.
We use the huggingface checkpoint\footnote{https://huggingface.co/microsoft/tapex-large-finetuned-tabfact} of the TAPEX-large version for the experiment.
Around 9\% of insights are filtered out during the Factuality Verification process. 

Subsequently, we apply another strategy called importance scoring (Algorithm~\ref{alg:insight_importance_scoring}) for the processed examples.
We adopt Llama-2-7b-chat~\citep{touvron2023llama} as summarizer $\theta$ to iteratively evaluate the impact of each ablated insight. 
In detail, we perform zero-shot inference conditioned on each ablated set of insights $\mathcal{\tilde I}$, since the model without prior knowledge of the table in the training data reacts more sensitively to each ablation. 
For similarity measure, we employ SBERT\footnote{ https://huggingface.co/sentence-transformers/stsb-xlm-r-multilingual} to compute the similarity between the generated summary and reference summary.

\begin{figure}[!t]
    \centering
    \includegraphics[width=1\columnwidth]{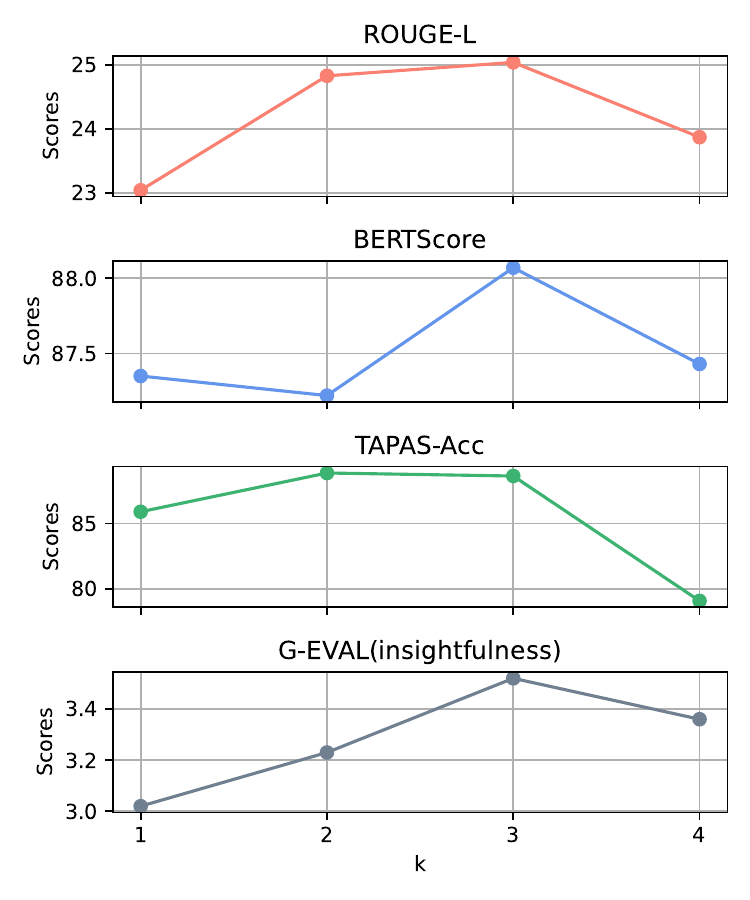}
    \caption{Summarization results with different $k$ 
    }
    \label{fig:ablation_k} 
\end{figure}

After the scoring process, we select top-k insights from each aspect. 
We set $k$ to 3 in our experiments. To understand the effect of $k$, we perform an ablation study using different numbers of $k$ to prune the insights.
Specifically, we adjust the number of insights in the training set in the range of 1 to 4 and train different reasoners for each $k$ to find the optimal $k$.
The results are shown in figure ~\ref{fig:ablation_k}.

In the end, we construct the pruned training set $D’$ which is an extended version of end-to-end table summarization training corpora $D$.
We provide the statistics of the generated dataset in Table~\ref{tab:data_stat_dprime}, and the filtered-out examples in Table~\ref{tab:case_pruned}.

\newcolumntype{L}{>{\raggedright\arraybackslash}p{0.2\columnwidth}}
\newcolumntype{C}{>{\centering}p{0.1\columnwidth}}


\begin{table}[!ht]
\centering
\resizebox{1.\columnwidth}{!}{
\begin{tabular}{c|c|cc|c|c}
\toprule
 \#Table & \#Aspect $\mathcal{A}$ &\#Question, \#Evidence, \#Insight ($\mathcal{Q},\mathcal{E},\mathcal{I} $) \\ \midrule
 2,054  & 9,207 & 27,621  \\

\bottomrule

\end{tabular}
}
\caption{Statistics of $D'$ used to train the \textsc{QtP} Reasoner.}
\label{tab:data_stat_dprime}
\end{table}
\subsection{\textsc{QtP} Reasoner Training Details}
\label{sec:QTP_training_detail}
We train \textsc{QtP} Reasoner with $D’$ on top of Llama-2-7b-chat~\citep{touvron2023llama} model with two different instruction tunning tasks.
Specifically, we randomly shuffle the instances from the aspect-focused question generation task and evidence-focused insight generation task, then jointly train them for a single model. 
To efficiently finetune the model, we adopt 4-bit quantized QLoRA and set the parameters as $r=64$, $\alpha = 16$. 
We use a constant learning rate schedule set at 2e-4, and train with the batch size of 4 on a single NVIDIA RTX A6000 GPU. The inference of the model is conducted using \texttt{vLLM}
framework\footnote{All open-source LLM inference in our experiments are conducted using the \texttt{vLLM}.}~\citep{10.1145/3600006.3613165}

\subsection{Dataset Details}
\label{sec:dataset_detail}
\paragraph{\textsc{InsTaSumm} Construction} To gain a better tesbed for evaluating the insightful summarization performance, we build a refined version of an existing dataset, namely \textsc{InsTaSumm}. We adopt QTSumm~\citep{zhao2023qtsumm} as a source dataset to construct \textsc{InsTaSumm}. 
We first collect all the tables and query-focused summaries in the train and test split of QTSumm, then aggregate it to a complete form of paragraph by prompting GPT-4 to verbalize it into a more fluent form. 
Table~\ref{tab:data_stat} shows the statistics of \textsc{InsTaSumm}.

\paragraph{SciGEN}
We choose SciGEN as an out-of-domain dataset to evaluate the generalization performance of \textsc{QtP} and other reasoner baselines.
We use the test split of the medium setting for the experiments. 

\paragraph{Dataset Statistics}
We provide dataset statistics of \textsc{InsTaSumm} and SciGEN in Table~\ref{tab:data_stat}. 
We report the number of table instances (\#Table) , average token length for each target summary (Avg.sum\_len), and the average length of columns and rows in a single table (Avg.tab\_len). 

\newcolumntype{L}{>{\raggedright\arraybackslash}p{0.2\columnwidth}}
\newcolumntype{C}{>{\centering}p{0.1\columnwidth}}


\begin{table}[!ht]
\centering
\resizebox{1.\columnwidth}{!}{
\begin{tabular}{c|c|cc|c|c}
\toprule
Dataset & Domain & Split &\#Table & Avg.sum\_len & Avg.tab\_len  \\ \midrule
\multirow{2}{*}{\textsc{InsTaSumm}} & \multirow{2}{*}{open}  & train & 2,054  & \multirow{2}{*}{161.9} &  \#col: 6.6\\
    &   &    test  & 440    &         & \#row: 11.71  \\
\midrule

\multirow{2}{*}{SciGEN} & \multirow{2}{*}{scientific}  & train & 13,607 & \multirow{2}{*}{115.3} &  \#col: 6.0\\
&   &     test   & 1,038     &     &  \#row: 7.64 \\ 

\bottomrule

\end{tabular}
}
\caption{Dataset Statistics of \textsc{InsTaSumm} and SciGEN.}
\label{tab:data_stat}
\end{table}

\subsection{Table Summarizer}
\label{sec:Table_summarizer}
We consider two different table summarizers (\textit{i.e.}, fine-tuned summarizer and zero-shot summarizer) for our experiments. 
For both scenarios, we provide the knowledge generated by \textsc{QtP} Reasoner as an additional input, concatenated to the serialized input table. 
For fine-tuned summarizers, we augment the input during both the training and inference phases, while for zero-shot summarizers, we augment the input only during the inference phase. 

\paragraph{Fine-tuned Summarizer}
We adopt two different open-source models, ReasTAP and Llama-2-7b-chat. 
\begin{itemize}[leftmargin=*,topsep=2pt,itemsep=2pt,parsep=0pt]
\item \textbf{ReasTAP}: 
ReasTAP~\citep{zhao-etal-2022-reastap} is a BART-based table-to-text model, pre-trained with synthetic table question and answering corpus.
we use the official implementation of ReasTAP-large version from official GitHub\footnote{https://github.com/Yale-LILY/ReasTAP} repository
\item \textbf{Llama-2-7b-chat}: 
Llama-2-7b-chat\footnote{https://huggingface.co/meta-llama/Llama-2-7b-hf}~\citep{touvron2023llama} is specifically optimized for conversational contexts with the instruction tuning on the top of Llama-2. 
\end{itemize}

\paragraph{Zero-shot Summarizer}
For zero-shot evaluation, we employ both open-source and closed-source LLM for the experiments. We adopt two large-scale models, GPT-3.5-turbo and Mistral-7b. 
\begin{itemize}[leftmargin=*,topsep=2pt,itemsep=2pt,parsep=0pt]
\item \textbf{GPT-3.5-turbo}: 
GPT-3.5-turbo~\citep{openai2023chatgpt} is an instruction-tuned chat LLM with 175B parameters. It stands as a prominent closed-source model renowned for its generalization ability in diverse NLP tasks. 
We use \texttt{GPT-3.5-turbo-0125} version API provided by OPENAI. 
\item \textbf{Mistral-7b}: 
Mistral\footnote{https://huggingface.co/mistralai/Mistral-7B-v0.1}~\citep{jiang2023mistral} is an opensource LLM that outperforms Llama-2-13B in diverse evaluated NLP benchmarks. 
\end{itemize}

\subsection{Baseline Resaoner Models}
\label{sec:baselines_detail}
To evaluate how knowledge affects the performance of summarization, we compare \textsc{QtP} Reasoner with other knowledge-augmented baselines that generate knowledge with two different types of reasoning \textit{i.e.,} step-by-step reasoning(CoT, Plan-and-Solve) and symbolic reasoning(Logical Type, SQL). 

For a fair comparison, we implement all baseline knowledge reasoners with the same backbone model as \textsc{QtP} Reasoner.
All reasoners are a student model trained on distilled knowledge which is generated by the teacher-LLM(GPT-3.5-turbo). We prompted LLM to generate training data for each reasoner from the reference summary and the input table with a 1-shot demonstration.
Same from Section ~\ref{sec:method}, the teacher model generates implicit knowledge from the table with different variants of reasoning strategy.

\paragraph{CoT Reasoner}
    
We first adopt Chain-of-Thought~\citep{NEURIPS2022_zeroshotcot} as the step-by-step reasoning strategy to generate the knowledge from the table. 
Specifically, the LLM is evoked to generate the reasoning step for each implicit knowledge with \textit{“Let’s think step by step”} prompt before the knowledge generation. 
    
\paragraph{Plan-and-Solve Reasoner}
We then adopt Plan-and-Solve~\citep{wang2023planandsolve} for the variation of step-by-step reasoning, where the high-level plan is first generated to solve the knowledge generation task, and the final knowledge is generated according to the plan with step-by-step reasoning.

\paragraph{Logical Type Reasoner}
We adopt Logical Type Reasoner for the symbolic reasoning-based knowledge model baseline. 
Logical Types~\citep{perlitz2023diversity,zhao-etal-2023-investigating} are widely used schemes in recent table-to-text literature that categorizes several logical table operations to search the information on the table cells. 

To apply the logical type in the process of knowledge generation, we adopt 9 predefined logical types (Negation, Superlative, Count, Comparative, Ordinal, Unique, All, and Surface) following ~\citet{zhao-etal-2023-investigating}, to sample the knowledge from the table by using each type as a control for each knowledge generation.
As simply providing all 9 types of knowledge from the table could be not helpful in generating the required knowledge for each reference summary, we first let the LLM choose the logical type that should be used to generates the reference summary.  
Then for each selected types, the model sequentially generate the corresponding table knowledge.

\paragraph{SQL Resaoner}
We adopt another symbolic reasoning baseline called SQL Reasoner, which uses SQL query for the intermediate control of the knowledge generation. 
Recent works~\citep{cheng2022binding, liu2022tapex, ye2023large, liu-etal-2022-plog, zhao-etal-2023-loft} in diverse table-based tasks have demonstrated that adopting the executable programs such as SQL or Logical Form shows remarkable performance improvement in table-related tasks. 
This is attributed to the complex nature of the structured table data, where logical programs can serve as faithful control for the information searching from the structured cells. 

Therefore, we follow the concept of ~\citep{ye2023large,liu-etal-2022-plog,zhao-etal-2023-loft} and let the LLM generate the implicit knowledge along with the SQL query for each knowledge. With the sequential prediction of each SQL query and knowledge, the model can be controlled to generate more faithful knoweldge~\citep{liu-etal-2022-plog} that contains table-related logical reasoning in the generated description.

\subsection{Evaluation Details}
\label{sec:evaluation_detail}
We evaluate the performance of \textsc{QtP} and the baselines with both automatic evaluation and human evaluation. 
With automatic evaluation, we assess the quality of the generated summary, while with human evaluation, we assess the quality of generated knowledge from \textsc{QtP} Reasoner and the baseline reasoner models.

\paragraph{Automatic Evaluation}
We evaluate the performance of summarization with four different perspectives: (1) Surface-level, (2) Faithfulness-level, (3) Insightfulness-level, and (4) Pairwise quality comparison.

\begin{itemize}
[leftmargin=*,topsep=2pt,itemsep=2pt,parsep=0pt]
\item \textbf{BLEU}: BLEU~\citep{10.3115/1073083.1073135} calculates the geometric mean of the precision over the n-grams of the output text. We utilized SacreBLEU~\citep{post-2018-call} to ensure consistent and reproducible BLEU scores.

\item \textbf{ROUGE}: ROUGE~\citep{lin-hovy-2003-automatic} evaluates word overlap between the candidate and reference summaries. We provided the F1 score for ROUGE-L, which considers the longest common subsequences.

\item \textbf{METEOR}: METEOR~\citep{banerjee-lavie-2005-meteor} focuses on a generalized concept of unigram matching between machine-generated translations and human reference translations.

\item \textbf{BERTScore}: BERTScore~\citep{Zhang*2020BERTScore:} measures the similarity between the reference and the generated summary by using contextual word embeddings.

\item \textbf{A3CU}: A3CU~\citep{liu-etal-2023-towards-interpretable} is an interpretable summarization evaluation system that aligns well with human judgments. It directly computes the similarity between texts without extracting atomic content units (ACUs) and uses the F1 score for evaluation.

\item \textbf{TAPAS-Acc}: TAPAS-Acc ~\citep{liu-etal-2022-plog} is a reference-free metric that leverages TAPAS~\citep{herzig-etal-2020-tapas} fine-tuned on the TabFact~\citep{Chen2020TabFact:} dataset to assess the faithfulness of the generated content.

\item \textbf{GPT4-Acc}: Following ~\citet{zhao-etal-2023-investigating}, we assess the faithfulness of generated using the GPT-4 as the backbone. It shows a better correlation with human judgments than the TAPAS-Acc.

\item \textbf{Insightfulness(G-EVAL)}: For insightfulness evaluation in Table ~\ref{tab:main_table}, we adopt the G-EVAL approach to assess the insightfulness of each summary. We use a 5-point Likert scale to score each summary and report the average score. We provide an evaluation prompt in Table ~\ref{tab:prompt_insightfulness}. 

\item \textbf{Pairwise Comparison}: For pairwise comparison in Figure ~\ref{fig:pairwise_summary}, we use GPT-4 to evaluate the summary quality in diverse criteria. Specifically, GPT-4 is prompted to choose the better quality summary for each criterion among two candidates. We provide the prompt used in the evaluation in Table~\ref{tab:prompt_pairwise_comprehensive},~\ref{tab:prompt_pairwise_informative} and~\ref{tab:prompt_pairwise_natural}. We adopt the following three criteria. 
\begin{itemize}
\item \textbf{ \textit{Natural}}: the extent how naturally the information is conveyed, reflecting an easy and relaxed use of language that is clear and correct.

\item \textbf{ \textit{Comprehensive}}: the extent to which the summary covers all the essential and important information presented in the source table.

\item \textbf{ \textit{Informative}}: the extent to which the summary provides clear, accurate, and relevant information derived from the source table, contributing effectively to the understanding of the table.
\end{itemize}
\end{itemize}

\begin{figure*}[!ht]
    \centering
    \includegraphics[width=1\textwidth]{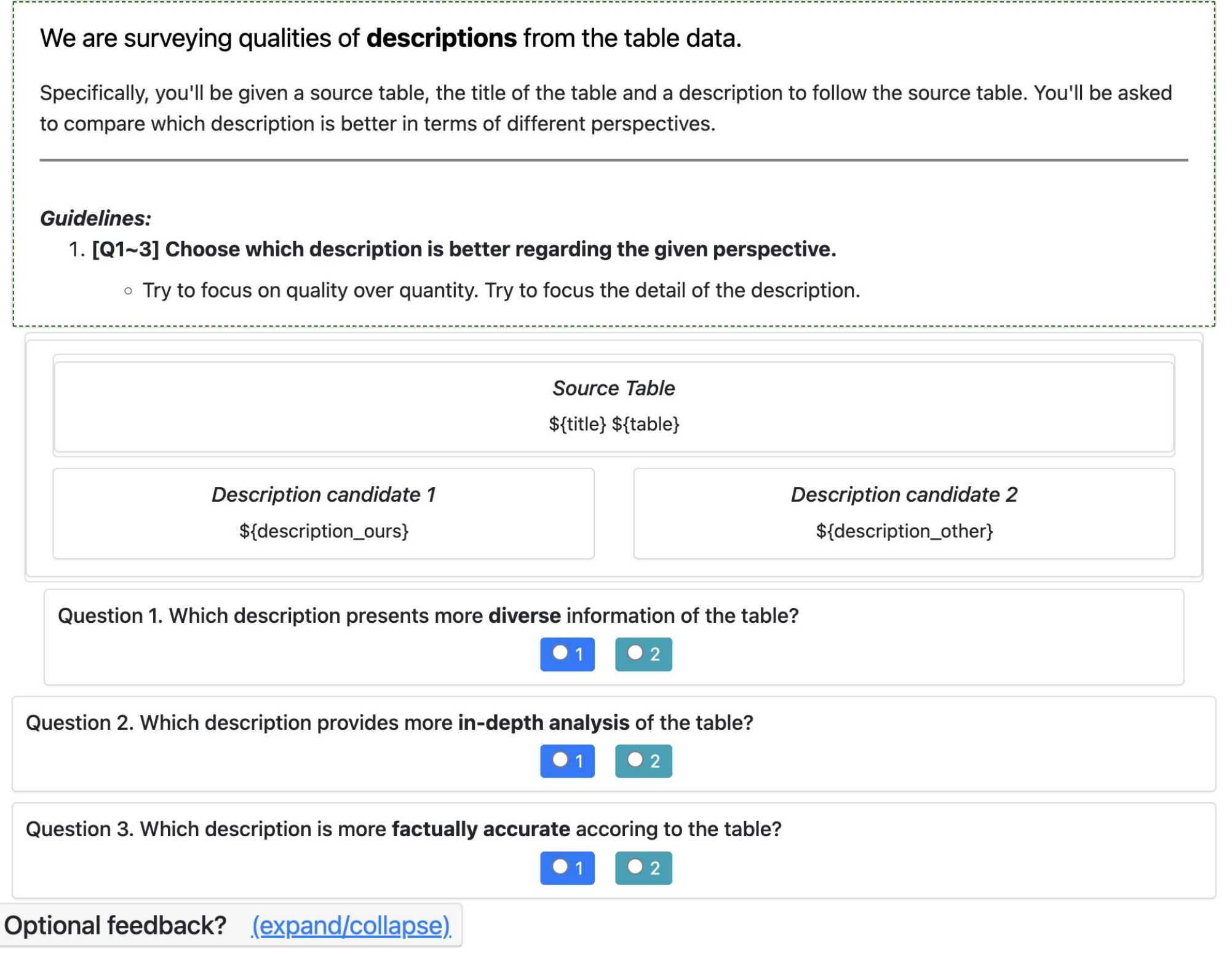}
    \caption{Annotator interface of human evaluation on reasoner generated knowledge quality 
    }
    \label{fig: human_eval_interface} ~
\end{figure*}

\paragraph{Human Evaluation}
We assess the quality of generated knowledge with the human evaluation by comparing the output of \textsc{QtP} Reasoner and those from other reasoner baselines via Amazon Mechanical Turk (AMT). 
We show the interface for the evaluation in Figure~\ref{fig: human_eval_interface}.
We ask three human judges to compare the quality of knowledge based on the following three criteria:
\begin{itemize}
[leftmargin=*,topsep=2pt,itemsep=2pt,parsep=0pt]
\item\textbf{\textit{Diverse}:} Which knowledge presents more diverse information from the table?

\item\textbf{\textit{Insightful}:} Which knowledge provides more in-depth analysis of the table?

\item\textbf{\textit{Faithful}:} Which knowledge is more accurate according to the table?
\end{itemize}

\subsection{Case Study}
We select representative examples of \textsc{QtP} and present in Table ~\ref{tab:case_ours_whole}  to~\ref{tab:case_pruned}
\label{sec:case_study_appendix}

\begin{table*}
    \small
    \centering
    \resizebox{1\textwidth}{!}{
        \begin{tabular}{p{0.18\textwidth}p{0.8\textwidth}>{\centering\arraybackslash}p{0.02\textwidth}}
            \toprule
            \multicolumn{2}{l}{\textbf{Example prediction of \textsc{QtP} from \textsc{InsTaSumm}}} \\
            \midrule
            Input Table & {title : 1990 - 91 Manchester United F.C. Season 
            
            col : Date | Opponents | H / A | Result F - A | Attendance 

            row 1 : 1 August 1990 | Bury | A | 0 - 0 | 7162 
            
            row 2 : 3 August 1990 | Cork City | A | 0 - 0 | 8000 
            
            row 3 : 5 August 1990 | Waterford United | A | 4 - 0 | 4750 
            
            row 4 : 8 August 1990 | Derry City | A | 1 - 1 | 9710 
            
            row 5 : 11 August 1990 | Irish League | N | 3 - 0 | 10037
            
            row 6 : 13 August 1990 | Bohemians | A | 3 - 0 | 13878
            
            row 7 : 15 August 1990 | Rangers | A | 1 - 0 | 31818
            
            row 8 : 20 November 1990 | Celtic | H | 1 - 3 | 41658} & \\
            \midrule
            Generated Knowledge             &  

            \textcolor{magenta}{\textbf{Aspect-a}}: Stadium Attendance Trends
            
            \textbf{Question-a$_1$}: How did the attendance numbers at matches held at Old Trafford evolve throughout the season?
            \textbf{Evidence-a$_1$}: The relevant columns and rows for the Question is col(Date, Attendance), row(5, 8). 
            \textcolor{teal}{\textbf{Insight-a$_1$}}: The attendance numbers at matches held at Old Trafford peaked at 41,658 for the game against Celtic on November 20, 1990. There were also matches with lower attendance figures, such as the match against the Irish League on August 11, 1990.
            
            \textbf{Question-a$_2$}: What were the factors that influenced the fluctuations in attendance at home matches?
            \textbf{Evidence-a$_2$}: The relevant columns and rows for the Question is col(Date, Opponents, Attendance), row(8)
            \textcolor{teal}{\textbf{Insight-a$_2$}}: The fluctuations in attendance at home matches could be influenced by various factors such as the opponents, the significance of the match, and the overall team performance.
            
            \textbf{Question-a$_3$}: How did the attendance at home matches reflect the fan sentiment or support for Manchester United during that period?
            \textbf{Evidence-a$_3$}: The relevant columns and rows for the Question is col(Date, Opponents, Attendance), row(8)
            \textcolor{teal}{\textbf{Insight-a$_3$}}: The attendance at home matches reflected varying levels of fan sentiment or support, with the highest attendance recorded in the game against Celtic, which ended in a 1-3 loss for Manchester United.
            \par\vspace{1em} 
            
            \textcolor{magenta}{\textbf{Aspect-b}}: Support from Fans
            
            \textbf{Question-b$_1$}: Were there any events or milestones during the 1990-91 season that influenced attendance numbers?
            \textbf{Evidence-b$_1$}: The relevant columns and rows for the Question is col(Date, Opponents, Attendance), row(8). 
            \textcolor{teal}{\textbf{Insight-b$_1$}}: The 20 November 1990 match against Celtic at Old Trafford was a significant turnaround point for United, marking their first loss in the league and garnering the highest attendance for a single game during the season.
            
            \textbf{Question-b$_2$}: How did Manchester United's fan base and support impact their overall performance and results during the season?
            \textbf{Evidence-b$_2$}: The relevant columns and rows for the Question is col(Opponents, Attendance), row(8).
            \textcolor{teal}{\textbf{Insight-b$_2$}}: Manchester United's fan base and support seemed to positively impact their overall performance and results during the season, as evidenced by the high attendance numbers and successful matches, such as the one against Celtic.
            
            \textbf{Question-b$_3$}: Were there any correlations between Manchester United's match outcomes and the attendance numbers at Old Trafford?
            \textbf{Evidence-b$_3$}: The relevant columns and rows for the Question is col(Date, Opponents, Result F - A, Attendance), row(8)
            \textcolor{teal}{\textbf{Insight-b$_3$}}: Manchester United faced defeat against Celtic but still managed to secure a high attendance of 41,658, indicating strong fan support regardless of match outcomes.
            \par\vspace{1em}
            
            \textcolor{magenta}{\textbf{Aspect-c}}: Results and Performance
            
            \textbf{Question-c$_1$}: What were the key highlights of Manchester United's performance against Celtic in terms of results and attendance?
            \textbf{Evidence-c$_1$}: The relevant columns and rows for the Question is col(Date, Opponents, Attendance), row(8)
            \textcolor{teal}{\textbf{Insight-c$_1$}}: Manchester United lost 1-3 to Celtic on November 20, 1990, with an attendance of 41,658.
            
            \textbf{Question-c$_2$}: Were there any specific factors that influenced the attendance trends at Manchester United's home games?
            \textbf{Evidence-c$_2$}: The relevant columns and rows for the Question is col(Opponents, Attendance), row(6, 8).
            \textcolor{teal}{\textbf{Insight-c$_2$}}: The high attendance at home games could be attributed to significant opponents like Bohemians and Celtic.
            
            \textbf{Question-c$_3$}: Did Manchester United face any challenges in maintaining consistent attendance numbers throughout the season?
            \textbf{Evidence-c$_3$}: The relevant columns and rows for the Question is col(Date, Attendance), row(1,2,3,4,5,6,7,8).
            \textcolor{teal}{\textbf{Insight-c$_3$}}: Manchester United faced fluctuations in attendance numbers throughout the season, with varying figures for different matches.
\\
            \midrule
            Reference Summary      & {
                The table detailing the 1990-91 Manchester United F.C. season provides a snapshot of key match details and attendance figures during that period. Notably, Manchester United faced Celtic on November 20, 1990, resulting in a 1-3 victory for Celtic, with an attendance of 41,658. Additionally, Manchester United played a match against an Irish League team on August 11, 1990, securing a 3-0 win with an attendance of 10,037. These records offer insights into the team's performance and fan engagement during the specified season, showcasing both victories and attendance numbers for the highlighted matches.
            } & \\
            \midrule
            Generated Summary      & {
                During the 1990-91 season, Manchester United had a strong performance in their away games, securing notable victories against clubs like Waterford United, Celtic, and Bohemians. The high attendance at these away matches reflected the strong support from their fan base. However, their home game against Celtic on November 20, 1990, marked a significant turnaround point as it resulted in their first loss in the league, despite attracting the highest attendance of 41,658 for a single game during the season. The fluctuating attendance at home matches seemed to be influenced by factors such as the opponents, match significance, and overall team performance. Despite facing defeat against Celtic, Manchester United continued to receive strong fan support, as evidenced by the high attendance numbers.
            } & \\


            \bottomrule
        \end{tabular}
    }
    \caption{Example of generated knowledge on InsTaSumm testset using GPT-3.5 as backbone summarizer.}
    \label{tab:case_ours_whole}
\end{table*}


\begin{table*}
    \small
    \centering
    \resizebox{1\textwidth}{!}{
        \begin{tabular}{p{0.18\textwidth}p{0.8\textwidth}>{\centering\arraybackslash}p{0.02\textwidth}}
            \toprule
            \multicolumn{2}{l}{\textbf{Example prediction of \textsc{QtP} from \textsc{SciGEN}}} \\
            \midrule
            Input Table & {title(caption) : Table 4: All experiment results of transfer learning and abbreviation expansion (top-2 scores marked as bold). MedNLI (expanded) denotes MedNLI with abbreviation expansion.
            
            col : Dataset | BERT dev | BERT test | BioBERT dev | BioBERT test
            
            row 1 : MedNLI | 79.56 | 77.49 | 82.15 | 79.04
            
            row 2 : MNLI (M) | 83.52 | - | 81.23 | -
            
            
            
            
            ...
            
            
            row 8 : MedNLI (expanded) | 79.13 | 77.07 | [BOLD] 83.87 | 79.95
            
            row 9 : S → M → MedNLI (expanded) | [BOLD] 82.15 | [BOLD] 79.95 | 83.08 | [BOLD] 81.85} & \\
            \midrule
            Generated Knowledge             &  

            \textcolor{magenta}{\textbf{Aspect-a}}: Interdataset Generalizability
            
            \textbf{Question-a$_1$}:Are there any instances where models from one dataset were able to transfer well to another dataset, indicating strong interdataset generalizability?
            \textbf{Evidence-a$_1$}: The relevant columns and rows for the Question is col(BERT dev), row(1, 2).
            \textcolor{teal}{\textbf{Insight-a$_1$}}: BERT models from the MedNLI dataset were able to transfer well to the MNLI (M) dataset, with an increase in performance on the MNLI (M) dataset.
            
            \textbf{Question-a$_2$}: What factors contribute to the differences in model performance based on the dataset they were fine-tuned on?
            \textbf{Evidence-a$_2$}: The relevant columns and rows for the Question is col(BERT dev, BERT test, BioBERT dev, BioBERT test), row(1, 3, 4, 5, 6, 7, 8, 9). 
            \textcolor{teal}{\textbf{Insight-a$_2$}}: The choice of dataset plays a crucial role in determining the performance of the model.
            
            \textbf{Question-a$_3$}: What implications do the varying scores have on the practical applications of models trained on different NLI datasets?
            \textbf{Evidence-a$_3$}: The relevant columns and rows for the Question is col(BERT dev, BERT test, BioBERT dev, BioBERT test), row(1, 3, 4, 5, 6, 7, 8, 9). 
            \textcolor{teal}{\textbf{Insight-a$_3$}}: The varying scores suggest that models trained on different NLI datasets have distinct strengths and potential applications in various NLP tasks requiring NLI knowledge.
            \par\vspace{1em} 
            
            
            
            
            \textcolor{magenta}{\textbf{Aspect-b}}: Transfer Learning Effectiveness
            
            \textbf{Question-b$_1$}: How effective is transfer learning across different NLI datasets based on the provided scores?
            \textbf{Evidence-b$_1$}: The relevant columns and rows for the Question is col(BioBERT dev, BioBERT test), row(1, 3, 4, 5, 6, 7, 9). 
            \textcolor{teal}{\textbf{Insight-b$_1$}}: Transfer learning from BioBERT achieves the highest scores on both MedNLI and SNLI datasets, with slightly lower scores on MNLI (M).
            
            \textbf{Question-b$_2$}: What challenges or advantages exist in transferring models pre-trained on MedNLI to other NLI datasets?
            \textbf{Evidence-b$_2$}: The relevant columns and rows for the Question is col(all), row(1,2,3,4,5,6,7, 8).
            \textcolor{teal}{\textbf{Insight-b$_2$}}: Transferring models pre-trained on MedNLI to other NLI datasets presents challenges in generalization and adaptation due todataset-specific recurrent patterns. However, advantages include shared vocabulary and concepts, facilitating initial model alignment.
            
            \textbf{Question-b$_3$}: How does the top-2 performance metric reflect the overall feasibility and impact of transfer learning in the NLI domain?
            \textbf{Evidence-b$_3$}: The relevant columns and rows for the Question is col(BERT dev, BERT test, BioBERT dev, BioBERT test), row(4, 7, 9). 
            \textcolor{teal}{\textbf{Insight-b$_3$}}: The top-2 performance metric reflects the success and capability of transfer learning in the NLI domain, with models trained on different datasets achieving high scores across various tasks.
            \par\vspace{1em}
            
            \textcolor{magenta}{\textbf{Aspect-c}}: Cross-Dataset Transferability
            
            \textbf{Question-c$_1$}: How does the top-2 performance metric evaluate the performance of models across different datasets?
            \textbf{Evidence-c$_1$}:  col(BERT dev, BERT test, BioBERT dev, BioBERT test), row(7, 9)
            \textcolor{teal}{\textbf{Insight-c$_1$}}: The top-2 performance metric evaluates the performance of models across different datasets by highlighting the top-scoring examples in each expansion scenario, indicating the highest-performing models for each dataset combination.
            
            \textbf{Question-c$_2$}: What factors contribute to the differences in performance of models across datasets, as depicted in the table?
            \textbf{Evidence-c$_2$}: The relevant columns and rows for the Question is col(BERT dev, BERT test, BioBERT dev, BioBERT test), row(1, 3, 4, 5, 6, 7, 8, 9)
            \textcolor{teal}{\textbf{Insight-c$_2$}}: The differences in performance of models across datasets could be attributed to variations in data and the specific challenges each dataset presents.
            
            \textbf{Question-c$_3$}: How does transferring models from one dataset to another impact the overall performance and applicability of the models in the NLI domain?
            \textbf{Evidence-c$_3$}: The relevant columns and rows for the Question is col(BERT dev, BERT test, BioBERT dev, BioBERT test), row(4, 7, 9).
            \textcolor{teal}{\textbf{Insight-c$_3$}}: Transferring models from one dataset to another generally results in an improvement in performance, indicating the potential benefits of transfer learning in the NLI domain.
\\
            \midrule
            Reference Summary      & {
                We conduct transfer learning on four different combinations of MedNLI, SNLI, and MNLI as it shown in the table 4 (line 4 to 7) and also add the results of general domain tasks (MNLI, SNLI) for comparison. BERT performs better on tasks in the general domain while BioBERT performs better on MedNLI which is in the clinical domain. positive transfer occurs on MedNLI. even though BioBERT is finetuned on general domain tasks before MedNLI, transfer learning shows better results than that fine-tuned on MedNLI directly. ... the accuracy of MNLI and SNLI on BioBERT is lower than the accuracy on BERT. The best combination is SNLI → MNLI → MedNLI on BioBERT. MedNLI (expanded) shows better performance than MedNLI on BioBERT while MedNLI works better on BERT (see table 4)...
            } & \\
            \midrule
            Generated Summary      & {
                The table presents the experiment results of transfer learning and abbreviation expansion, with top-2 scores marked in bold. The performance of models on MedNLI (expanded) shows a slight decrease compared to the original MedNLI, with a decrease in test scores from 79.95 to 79.04. The BERT base model, bi-directional learning, ... Transfer learning from BioBERT achieves the highest scores on both MedNLI and SNLI datasets, with slightly lower scores on MNLI (M). Models trained on different NLI datasets have distinct strengths and potential applications in various NLP tasks requiring NLI knowledge. Transferring models from one dataset to another generally results in an improvement in performance, indicating the potential benefits of transfer learning in the NLI domain.
            } & \\


            \bottomrule
        \end{tabular}
    }
    \caption{Example of generated knowledge on SciGEN testset using GPT-3.5 as backbone summarizer.}
    \label{tab:case_ours_whole_scigen}
\end{table*}


\begin{table*}
    \small
    \centering
    \resizebox{1\textwidth}{!}{
        \begin{tabular}{p{0.18\textwidth}p{0.8\textwidth}>{\centering\arraybackslash}p{0.02\textwidth}}
            \toprule
            \multicolumn{2}{l}{\textbf{Example of Summary prediction from \textsc{InsTaSumm}}} \\
            \midrule
            Input Table & {title : 1990 - 91 Manchester United F.C. Season 
            
            col : Date | Opponents | H / A | Result F - A | Attendance 

            row 1 : 1 August 1990 | Bury | A | 0 - 0 | 7162 
            
            row 2 : 3 August 1990 | Cork City | A | 0 - 0 | 8000 
            
            row 3 : 5 August 1990 | Waterford United | A | 4 - 0 | 4750 
            
            row 4 : 8 August 1990 | Derry City | A | 1 - 1 | 9710 
            
            row 5 : 11 August 1990 | Irish League | N | 3 - 0 | 10037
            
            row 6 : 13 August 1990 | Bohemians | A | 3 - 0 | 13878
            
            row 7 : 15 August 1990 | Rangers | A | 1 - 0 | 31818
            
            row 8 : 20 November 1990 | Celtic | H | 1 - 3 | 41658} & \\

            \midrule
            Without Knowledge      & 
                The table includes the date of the match, the opponents, whether the match was home or away, the result in terms of goals scored and conceded, and the attendance for each match. The season started with a series of away matches, with varying results and attendance numbers. The table also includes a home match against Celtic, which resulted in a 1-3 loss with a high attendance of 41658.
             & \\
            
            \midrule
            + CoT Reasoner      & { In the 1990 season, the team played a series of matches, with the highest attendance of 41,658 spectators recorded at the match against Celtic on 15 August 1990 at Celtic Park. The match against Rangers on 15 August 1990 at a Football Association (FA) shared ground had an attendance of 31,818 spectators. Notably, all the away matches resulted in draws, with the team playing against Bury, Cork City, Waterford United, Derry City, Irish League, Bohemians, and Rangers, with results of 0-0, 0-0, 4-0, 1-1, 3-0, 3-0, and 1-0 respectively.
                
            } & \\
            \midrule
            + Plan-and-Solve Reasoner      & { During the 1990/91 season, Linfield F.C. had a mixed performance with three losses. The team played a series of matches, with a notable range in attendance numbers, from 13,378 to 42,500. The most significant attendance was recorded at 138,780 for their game against Rangers on 15 August 1990. The team had a variety of results, including draws and wins, with the most notable being a 5-0 victory against Irish League on 11 August 1990.
                
            } & \\

            \midrule
            + Logical Type Reasoner      & { During the 1990-1991 season, a total of 8 games were played by the team. The matches were evenly split between home and away games, with the highest attendance recorded at 31818 during the away game against Rangers. The team had a strong start with a 4-0 victory against Waterford United and a 3-0 win against Bohemians. However, they faced a 1-3 defeat at home against Celtic, who outscored them with a score difference of 2-3. Overall, the team consistently attracted a high attendance, with no game having less than 10,000 spectators.
                
            } & \\

            \midrule
            + SQL Reasoner      & { In August 1990, the football team played a series of matches, including away games against Bury, Cork City, Waterford United, Derry City, Bohemians, and Rangers, as well as a neutral game against the Irish League. The team achieved a 4-0 victory against Waterford United and a 3-0 victory against both the Irish League and Bohemians. The match against Rangers on August 15, 1990, had the highest attendance of 31,818 at Lansdowne Road Stadium. The team also played a home game against Celtic in November 1990, resulting in a 1-3 loss with an attendance of 41,658.
                
            } & \\

            \midrule
            \textbf{+ \textsc{QtP} Reasoner}     & { During the 1990-91 season, Manchester United had a strong performance in their away games, securing notable victories against clubs like Waterford United, Celtic, and Bohemians. The high attendance at these away matches reflected the strong support from their fan base. However, their home game against Celtic on November 20, 1990, marked a significant turnaround point as it resulted in their first loss in the league, despite attracting the highest attendance of 41,658 for a single game during the season. The fluctuating attendance at home matches seemed to be influenced by factors such as the opponents, match significance, and overall team performance. Despite facing defeat against Celtic, Manchester United continued to receive strong fan support, as evidenced by the high attendance numbers.
                
            } & \\
            \bottomrule
        \end{tabular}
    }
    \caption{Example of generated Summary on InsTaSumm testset using GPT-3.5 as backbone summarizer paired with different baselines.}
    \label{tab:case_summary_instasumm}
\end{table*}

\begin{table*}
    \small
    \centering
    \resizebox{1\textwidth}{!}{
        \begin{tabular}{p{0.18\textwidth}p{0.8\textwidth}>{\centering\arraybackslash}p{0.02\textwidth}}
            \toprule
            \multicolumn{2}{l}{\textbf{Example of Knowledge prediction from \textsc{InsTaSumm}}} \\
            \midrule
            Input Table & {title : 1990 - 91 Manchester United F.C. Season 
            
            col : Date | Opponents | H / A | Result F - A | Attendance 

            row 1 : 1 August 1990 | Bury | A | 0 - 0 | 7162 
            
            row 2 : 3 August 1990 | Cork City | A | 0 - 0 | 8000 
            
            row 3 : 5 August 1990 | Waterford United | A | 4 - 0 | 4750 
            
            row 4 : 8 August 1990 | Derry City | A | 1 - 1 | 9710 
            
            row 5 : 11 August 1990 | Irish League | N | 3 - 0 | 10037
            
            row 6 : 13 August 1990 | Bohemians | A | 3 - 0 | 13878
            
            row 7 : 15 August 1990 | Rangers | A | 1 - 0 | 31818
            
            row 8 : 20 November 1990 | Celtic | H | 1 - 3 | 41658} & \\

            \midrule
            CoT Reasoner      & { 
                                    $\bullet$ The match against Celtic on 20 November 1990 at Celtic Park had the highest attendance of 41,658 spectators.
                                    
                                    $\bullet$ The match against Rangers on 15 August 1990 at a Football Association (FA) shared ground had an attendance of 31,818 spectators. 
                                    
                                    $\bullet$ All the away matches ended in draws. The match against Cork City on 3 August 1990 ended in a draw with an attendance of 8,000 spectators.
                                    
                                    $\bullet$ The game against Waterford United on 5 August 1990 saw an attendance of 4,750, with a final score of 4-0.
                                    
                                    $\bullet$ On 8 August 1990, the match against Derry City ended in a 1-1 draw, attended by 9,710 fans.

            } & \\
            \midrule
            Plan-and-Solve Reasoner      & { 
            $\bullet$ Linfield F.C. had five losses during the 1990/91 season. There was a significant range in attendance numbers, from 13,878 to 47,500.
            
            $\bullet$ The most notable attendance was 138,780 for their game against Rangers on 15 August 1990. 
            
            $\bullet$ The match against the Irish League on 11 August 1990, held at a neutral venue, ended with a score of 3-0 and had 10,037 spectators.
            
            $\bullet$ A match against Bohemians on 13 August 1990, witnessed by 13,878 fans, concluded with a 3-0 result.
            
            $\bullet$ The game against Bury on 1 August 1990 ended in a 0-0 draw, attracting 7,162 spectators.
            
            } & \\

            \midrule
            Logical Type Reasoner      & { 
            $\bullet$ The total number of games played during the 1990-1991 season was 8. There were no games where the attendance was less than 10,000. 
            
            $\bullet$ The match against Rangers had the highest attendance of 318,180. Celtic outscored the opponents with a score difference of 2-3. 
            
            $\bullet$ The match against Celtic on 20 November 1990, which was held at home, ended in a 1-3 result, with the highest attendance of 41,658 spectators. 
            
            $\bullet$ On August 1, 1990, the match against Bury saw an attendance of 7,162 spectators.
            
            $\bullet$ The game against Bohemians on August 13, 1990, witnessed by 13,878 fans, recorded a significant attendance.

            } & \\

            \midrule
            SQL Reasoner      & { 
            $\bullet$ The match with the lowest attendance at Lansdowne Road Stadium was against Rangers on August 15, 1990. 
            
            $\bullet$ The match against Rangers on August 15, 1990 had the highest attendance of 318,180. The match against Celtic on November 20, 1990, held at Celtic Park, had the highest attendance of 41,350 spectators.

            $\bullet$ On August 15, 1990, the game against Rangers, played at Lansdowne Road Stadium, drew the largest crowd of 31,818 fans.
            
            $\bullet$ The lowest attendance of 8,000 was recorded at the match against Cork City on August 3, 1990.
            
            $\bullet$ The match at home against Celtic on November 20, 1990, drew a significant crowd of 41,156 spectators.
            
            $\bullet$ On August 15, 1990, the match against Rangers, which had the highest attendance of 31,818, was played at a shared ground.
                
            } & \\

            \midrule
            \textbf{\textsc{QtP} Reasoner}     & {
            
            $\bullet$ Manchester United's fan base and support seemed to positively impact their overall performance and results during the season, as evidenced by the high attendance numbers and successful matches, such as the one against Celtic.
            
            $\bullet$ Manchester United faced defeat against Celtic but still managed to secure a high attendance of 41,658, indicating strong fan support regardless of match outcomes.

            $\bullet$ The fluctuations in attendance at home matches could be influenced by various factors such as the opponents, the significance of the match, and the overall team performance., 
            
            $\bullet$ The attendance numbers at matches held at Old Trafford peaked at 41,658 for the game against Celtic on November 20, 1990. 
            
            $\bullet$ There were also matches with lower attendance figures, such as the match against the Irish League on August 11, 1990.', 
            
            $\bullet$ The attendance at home matches reflected varying levels of fan sentiment or support, with the highest attendance recorded in the game against Celtic, which ended in a 1-3 loss for Manchester United., 
            
            $\bullet$ Manchester United faced fluctuations in attendance numbers throughout the season, with varying figures for different matches.
                
            } & \\
            \bottomrule
        \end{tabular}
    }
    \caption{Example of generated knowledge on InsTaSumm testset using GPT-3.5 as backbone summarizer.}
    \label{tab:case_knowledge_instasumm}
\end{table*}

\begin{table*}
    \small
    \centering
    \resizebox{1\textwidth}{!}{
        \begin{tabular}{p{0.18\textwidth}p{0.8\textwidth}>{\centering\arraybackslash}p{0.02\textwidth}}
            \toprule
            \multicolumn{2}{l}{\textbf{Filtered-out training example in $D'$ after Knowledge Quality Enhancement}} \\
            \midrule
            Input Table & {title : 2008 in DREAM - Events list
                        
                        col : \# | Event Title | Date | Arena | Location | Attendees | Broadcast
                        
                        row 1 : 7 | Fields Dynamite!! 2008 | December 31, 2008 | Saitama Super Arena | Saitama, Saitama, Japan | 25,634 | Tokyo Broadcasting System; HDNet
                        
                        row 2 : 6 | Dream 6: Middleweight Grand Prix 2008 Final Round | September 23, 2008 | Saitama Super Arena | Saitama, Saitama, Japan | 20,929 | SkyPerfect; HDNet
                        
                        row 3 : 5 | Dream 5: Lightweight Grand Prix 2008 Final Round | July 21, 2008 | Osaka-jo Hall | Osaka, Osaka, Japan | 11,986 | SkyPerfect; HDNet
                        
                        row 4 : 4 | Dream 4: Middleweight Grand Prix 2008 Second Round | June 15, 2008 | Yokohama Arena | Yokohama, Kanagawa, Japan | 14,037 | SkyPerfect; HDNet
                        
                        row 5 : 3 | Dream 3: Lightweight Grand Prix 2008 Second Round | May 11, 2008 | Saitama Super Arena | Saitama, Saitama, Japan | 21,789 | SkyPerfect; HDNet
                        
                        row 6 : 2 | Dream 2: Middleweight Grand Prix 2008 First Round | April 29, 2008 | Saitama Super Arena | Saitama, Saitama, Japan | 21,397 | SkyPerfect; HDNet
                        
                        row 7 : 1 | Dream 1: Lightweight Grand Prix 2008 First Round | March 15, 2008 | Saitama Super Arena | Saitama, Saitama, Japan | 19,120 | Tokyo Broadcasting System; HDNet} & \\
            \midrule
            Mined \& Pruned Knowledge             &  

            \textcolor{magenta}{\textbf{Aspect-a}}: Audience Engagement
            \par\vspace{1em} 
            
            \textcolor{teal}{\textbf{Insight-a$_1$}}: The total attendance at DREAM events in 2008 was 125,522. 
            
            \textcolor{red}{\textit{(Factuality Verification filtered)}}
            \par\vspace{1em} 
            
            \textcolor{teal}{\textbf{Insight-a$_2$}}: The audience size varied based on the type of event or tournament, with some events attracting higher attendance figures than others.
            \par\vspace{1em} 
            
            \textcolor{teal}{\textbf{Insight-a$_3$}}: It was broadcasted by Tokyo Broadcasting System and HDNet, ensuring widespread coverage and viewer engagement both locally in Japan and internationally. 
            
            \textcolor{red}{\textit{(Importance Scoring filtered)}}
            \par\vspace{1em} 

            \textcolor{teal}{\textbf{Insight-a$_4$}}: The broadcast of events by different networks like Tokyo Broadcasting System and HDNet may have influenced the overall audience engagement with DREAM events in 2008. 
            \par\vspace{1em} 

            \textcolor{teal}{\textbf{Insight-a$_5$}}:The attendance figures, especially the high numbers at Saitama Super Arena events, reflect the strong interest and appeal of DREAM events to the public in 2008.
            \par\vspace{1em} 
            
            \textcolor{magenta}{\textbf{Aspect-b}}: Geographical Spread
            \par\vspace{1em} 
            
            \textcolor{teal}{\textbf{Insight-b$_1$}}: Three different venues hosted DREAM events in 2008: Saitama Super Arena in Saitama, Osaka-jo Hall in Osaka, and Yokohama Arena in Kanagawa. 
            
            \textcolor{red}{\textit{(Importance Scoring filtered)}}
            \par\vspace{1em} 
            
            \textcolor{teal}{\textbf{Insight-b$_2$}}: This global reach underscored the growing popularity of MMA and the organizational prowess of DREAM in staging high-profile events that captured the attention of diverse audiences across different continents. 
            
            \textcolor{red}{\textit{(Importance Scoring filtered)}}
            \par\vspace{1em} 
            
            \textcolor{teal}{\textbf{Insight-b$_3$}}:  The geographical spread of events did impact the overall audience engagement, with Saitama Super Arena attracting the highest attendance figures compared to Osaka-jo Hall and Yokohama Arena.
            \par\vspace{1em}

            \textcolor{teal}{\textbf{Insight-b$_4$}}: The Saitama Super Arena emerged as the most frequented location, with impressive attendance figures, possibly due to its popularity and draw compared to other venues like Osaka-jo Hall and Yokohama Arena
            \par\vspace{1em}

            \textcolor{teal}{\textbf{Insight-b$_5$}}: The attendance at events in Saitama, particularly at the Saitama Super Arena, was significantly higher compared to events in Osaka and Kanagawa, indicating varying levels of audience engagement across these locations. 

            \par\vspace{1em}

            ...
            
            & \\
            
            \midrule
            Reference Summary      & {
                The table "2008 in DREAM - Events list" provides a detailed overview of the DREAM events held in 2008, showcasing the distribution of events across different arenas and their respective attendance figures. Notably, the majority of events (four) took place at the Saitama Super Arena, attracting a total of 108,869 attendees, while two events were hosted in other arenas with a combined attendance of 26,023. This stark contrast in attendance between events at the Saitama Super Arena and other venues highlights the popularity and draw of the former location. Additionally, the geographical spread of DREAM events in 2008 encompassed three distinct venues: Saitama Super Arena in Saitama, Osaka-jo Hall in Osaka, and Yokohama Arena in Kanagawa. Among these, the Saitama Super Arena emerged as the most frequented location, with an impressive attendance of 25,634. Furthermore, the televised events in 2008, namely Fields Dynamite!! 2008 and Dream1:Lightweight Grand Prix 2008 First Round, broadcasted by the Tokyo Broadcasting System, attracted significant audiences of 25,634 and 19,120 attendees respectively. This data underscores the diverse event landscape of DREAM in 2008 and sheds light on the varying levels of audience engagement across different venues and broadcasts.
            } &\\
            \bottomrule
        \end{tabular}
    }
    \caption{Filtered-out example from pruned dataset $D'$ after the knowledge quality enhancement.}
    \label{tab:case_pruned}
\end{table*}


                
                




\begin{table*}
    \small
    \centering
    \begin{tabular}{p{14cm}}
    \toprule
    \textbf{Coarse-to-fine Knowledge Mining (Generate $\mathcal{A}, \mathcal{Q}$)} \\
    \midrule
\textcolor{teal}{\textbf{[Task Description]}}\\

Generate intermediate knowledge from the table for generating the target summary("Summary:"). The knowledge should contain (Coarse-level Aspects) and (Fine-level Questions for each Aspects). These knowledge should be a crucial cue to generate the target summary, but you should not explicitly state the target summary and also pretend that you don't know the target summary. Generate five Fine-level Questions for each Aspects. The questions must be the list of five different strings. Use Example 1 as reference, respond to Example2. 
    \\\\

    \textcolor{teal}{\textbf{[Example 1]}} \\
\textbf{Table:} \textit{title + serialized table}
\\
\textbf{Summary:} \textit{target summary}

\\
\textbf{Knowledge:} 
\\
\textit{(Coarse-level Aspect): aspect 1} 
\\
\textit{(Fine-level Questions):}
\\
\textit{Q1-1, Q1-2, ... Q1-5} 
\\
...
\\
\textit{(Coarse-level Aspect): aspect $n$} 
\\
\textit{(Fine-level Questions):}
\\
\textit{Q$n$-1, Q$n$-2, ... Q$n$-5}
\\\\

    \textbf{\textcolor{teal}{\textbf{[Example 2]}}} \\

    \textbf{Table:} \{table\}
\\
\textbf{Summary:} \{summary\}

\\
\textbf{Knowledge:}
    
\\ 
\bottomrule
    \end{tabular}
    \caption{The prompt used for Coarse-to-Fine Knowledge Mining (Aspect and Question).}
    \label{tab:crf_AQ}
\end{table*}


\begin{table*}
    \small
    \centering
    \begin{tabular}{p{14cm}}
    \toprule
    \textbf{Coarse-to-fine Knowledge Mining (Generate $\mathcal{E}, \mathcal{I}$)} \\
    \midrule
\textcolor{teal}{\textbf{[Task Description]}}\\

First, find answer for each given questions from the given table. Refer to the summary if it is needed. Then, select releavant column and rows from the table to answer the given questions from the table. Be careful not to omit any questions, answer all questions provided. 
Use Example 1 as reference, respond to Example2.     
    \\\\

    \textcolor{teal}{\textbf{[Example 1]}} \\
\textbf{Table:} \textit{serialized table}
\\
\textbf{Summary:} \textit{target summary}

\\
\textbf{Questions:} 
\\
\textit{Q-1, Q-2, ... Q-5} 
\\
\textbf{Question, Relevant column \& row , Answer triples:} \textit{(Q-1, E-1, I-1)}
\\\\

    \textbf{\textcolor{teal}{\textbf{[Example 2]}}} \\

    \textbf{Table:} \{table\}
\\
\textbf{Summary:} \{summary\}

\\
\textbf{Questions:} \{questions\}

\\
\textbf{Question, Relevant column \& row , Answer triples:}
    
\\
\bottomrule
    \end{tabular}
    \caption{The prompt used for Coarse-to-Fine Knowledge Mining (Evidence and Insight). }
    \label{tab:crf_EI}
\end{table*}

\begin{table*}
    \small
    \centering
    \begin{tabular}{p{14cm}}
    \toprule
    \textbf{Reasoner Training Instruction prompt (Aspect-focused Question Generation Task)} \\
    \midrule
\textcolor{teal}{\textbf{[Task Description]}}\\

Generate the intermediate knowledge for making the insightful summary of the table. First, the Knowledge should contain multiple (Coarse-level Aspect). Then for each single (Coarse-level Aspect), multiple (Fine-level Question) should be generated. These knowledge should be a crucial cue to generate the insightful summary.     
\\\\

\textbf{Table:} \{table\}
\\

\textbf{Response:}
    
\\
\bottomrule
    \end{tabular}
    \caption{The instruction prompt used for training reasoner in Aspect-Focused Question Generation Task }
    \label{tab:QtP_instruct_QG}
\end{table*}

\begin{table*}
    \small
    \centering
    \begin{tabular}{p{14cm}}
    \toprule
    \textbf{Reasoner Training Instruction prompt (Evidence-focused Insight Generation Task)} \\
    \midrule
\textcolor{teal}{\textbf{[Task Description]}}\\

Your task is find insights from the table by answering for the given Question. You will be provided a Table and Question. Before finding answer from the table, carefully look up the given table and pinpoint the relevant columns and rows as the evidence for the answer.    
\\\\

\textbf{Table:} \{table\}
\\

\textbf{Questions:} \{question\}
\\
\textbf{Response:}
    
\\
\bottomrule
    \end{tabular}
    \caption{The instruction prompt used for training reasoner in Evidence-Focused Insight Generation Task}
    \label{tab:QtP_instruct_IG}
\end{table*}

\begin{table*}
    \small
    \centering
    \begin{tabular}{p{14cm}}
    \toprule
    \textbf{Zero-shot Summariazation prompt} \\
    \midrule
\textcolor{teal}{\textbf{[Task Description]}}\\

Generate Summary of the given table based on the given Knowledge. You can refer to the Knowledge to make a complete paragraph-form summary.     
\\\\

\textbf{Table:} \{table\}
\\

\textbf{Knowledge:} \{generated \_knowledge\}
\\
\textbf{Summary:}
    
\\
\bottomrule
    \end{tabular}
    \caption{The prompt used for zero-shot summarization. We provide generated knowledge from \textsc{QtP} and other reasoner baselines as the additonal input.}
    \label{tab:prompt_zero-shot-SG}
\end{table*}

\begin{table*}
    \small
    \centering
    \begin{tabular}{p{14cm}}
    \toprule
    \textbf{Insightfulness-level Evaluation prompt (G-EVAL) } \\
    \midrule
\textcolor{teal}{\textbf{[Task Description]}}\\

You will be given one summary written for a table.
Your task it to rate the summary on one metric.
Please make sure you read and understand these instructions carefully. Please keep this document open while reviewing, and refer to it as needed.
\\\\  

Evaluation Criteria:  
\\
Analytical Depth (1-5) - the extent to which the summary provides insightful interpretations, explains the implications, and contextualizes the information from the source table.
\\\\

Evaluation Steps:
\\
1. Examine the source table thoroughly, identifying the main topic, all key data points, and potential implications or contexts that could be drawn from the data.
\\
2. Review the summary and assess its depth in analyzing the table. Determine if the summary offers deeper insights beyond the basic data, including interpretations of the implications, the significance of the information, and any contextual analyses.
\\
3. Assign a score for Analytical Depth on a scale of 1 to 5, where 1 indicates a minimal or superficial analysis with no significant insights beyond the direct data, and 5 indicates a highly insightful summary that effectively delves into the broader implications, contextual significances, and thorough interpretations of the information from the table.
\\\\

\textbf{Source Table:} \{table\}
\\

\textbf{Summary:} \{summary\}
\\
\textbf{Evaluation Form(scores ONLY):}
\\
\- \textbf{ Analytic Depth:}
\\
\bottomrule
    \end{tabular}
    \caption{The prompt for insightfulness-level evaluation. We adopt G-EVAL~\citep{liu2023geval} approach.}
    \label{tab:prompt_insightfulness}
\end{table*}


\begin{table*}
    \small
    \centering
    \begin{tabular}{p{14cm}}
    \toprule
    \textbf{Faithfulness-level summary evaluation (GPT4-Acc) } \\
    \midrule
\textcolor{teal}{\textbf{[Task Description]}}\\

Your task is to evaluate whether the given summary faithfully state the facts from the given source table. Read the table and check whether each single sentence in the summary is true or false. Think step-by-step to verify each sentence. Use Example1 as reference, respond to Example2. 

\\\\

\textcolor{teal}{\textbf{[Example 1]}}\\

\textbf{Table:} \textit{table}
\\

\textbf{Summary:} \textit{generated\_summary}

\\\\
\textbf{Evaluation:}
\\
(Sentence1): \textit{sent1}
\\
(explanation): \textit{explanation}
\\
(Verification): \textit {T/F}
\\

... 

(Sentence $n$): \textit{sent $n$}\\
(explanation): \textit{explanation}\\
(Verification): \textit {T/F}
\\\\

\textcolor{teal}{\textbf{[Example 2]}}\\

\textbf{Table:} \{table\}
\\
\textbf{Summary:} \{summary\}
\\\\
\textbf{Evaluation:}
\\
\bottomrule
    \end{tabular}
    \caption{The prompt for faithfulness-level summary evaluation.}
    \label{tab:prompt_faithfulness}
\end{table*}

\begin{table*}
    \small
    \centering
    \begin{tabular}{p{14cm}}
    \toprule
    \textbf{Pairwise Summary Quality Comparison (Criteria: Comprehensive) } \\
    \midrule
\textcolor{teal}{\textbf{[Task Description]}}\\

You will be given two different summary written for a table.
Your task it to rate each summary on one metric.
\\
Please make sure you read and understand these instructions carefully. Please keep this document open while reviewing, and refer to it as needed.
\\\\
Evaluation Criteria:
\\
Comprehensiveness - the extent to which the summary covers all the essential and important information presented in the source table.
\\\\
Evaluation Steps:
\\
1. Examine the source table thoroughly, identifying the main topic and all key points presented.
\\
2. Review the two different summary and compare each with the source table. Determine if the summary encapsulates all essential information without omitting any significant details.
\\
3. Assign two score for comprehensiveness on a scale of 1 to 5 for each summary, where 1 indicates a minimal coverage of information and 5 indicates a comprehensive summary that includes all essential details from the source table
\\
4. Choose the better summary in terms of the comprehensiveness score. 
\\\\
\textbf{Source Table:} \{table\}
\\\\

\textbf{Summary A:} \{summary\_A\}
\\\\
\textbf{Summary B:} \{summary\_B\}
\\\\
Generate Your assessment in the following format:
\\
\textbf{Better Summary Index: [Index]}(Use A for select Summary A and B for select Summary B)

\\
\bottomrule
    \end{tabular}
    \caption{The prompt for pairwise summary quality comparison (criteria: comprehensive) }
    \label{tab:prompt_pairwise_comprehensive}
\end{table*}


\begin{table*}
    \small
    \centering
    \begin{tabular}{p{14cm}}
    \toprule
    \textbf{Pairwise Summary Quality Comparison (Criteria: Informative) } \\
    \midrule
\textcolor{teal}{\textbf{[Task Description]}}\\

You will be given two different summary written for a table.
Your task it to rate each summary on one metric.
\\
Please make sure you read and understand these instructions carefully. Please keep this document open while reviewing, and refer to it as needed.
\\\\
Evaluation Criteria:
\\
Informativeness - the extent to which the summary provides clear, accurate, and relevant information derived from the source table, contributing effectively to the understanding of the table.
\\\\
Evaluation Steps:
\\
1. Carefully examine the source table to understand the main topic and capture all pertinent information.
\\
2. Review each of the two summaries to assess how effectively they convey the key information from the source table. Evaluate if the summaries provide clear and relevant information that aids in understanding the main points.
\\
3. Assign a score for informativeness on a scale of 1 to 5 for each summary, where 1 indicates that the summary provides minimal or unclear information, and 5 indicates that the summary offers clear, accurate, and relevant information enhancing understanding of the main topic.
\\
4. Determine and indicate which summary is better in terms of informativeness by comparing the assigned scores.
\\\\
\textbf{Source Table:} \{table\}
\\\\

\textbf{Summary A:} \{summary\_A\}
\\\\
\textbf{Summary B:} \{summary\_B\}
\\\\
Generate Your assessment in the following format:
\\
\textbf{Better Summary Index: [Index]}(Use A for select Summary A and B for select Summary B)

\\
\bottomrule
    \end{tabular}
    \caption{The prompt for pairwise summary quality comparison (criteria: informative) }
    \label{tab:prompt_pairwise_informative}
\end{table*}


\begin{table*}
    \small
    \centering
    \begin{tabular}{p{14cm}}
    \toprule
    \textbf{Pairwise Summary Quality Comparison (Criteria: Natural) } \\
    \midrule
\textcolor{teal}{\textbf{[Task Description]}}\\

You will be given two different summary written for a table.
Your task it to rate each summary on one metric.
\\
Please make sure you read and understand these instructions carefully. Please keep this document open while reviewing, and refer to it as needed.
\\\\
Evaluation Criteria:
\\
Naturalness - the extent to which the summary reads naturally, reflecting an easy and relaxed use of language that is clear and correct.
\\\\
Evaluation Steps:
\\
1. Thoroughly understand the source table, including its data and context.
\\
2. Review each of the two summaries, assessing how naturally the information is conveyed. Focus on the use of language, including syntax, semantics, and overall coherence.
\\
3. Assign a naturalness score from 1 to 5 for each summary. A score of 1 indicates that the summary is unnatural or difficult to understand, while a score of 5 suggests that the summary reads clearly and smoothly, with a natural flow.
\\
4. Compare the scores to determine which summary best achieves naturalness in its language and presentation.
\\\\
\textbf{Source Table:} \{table\}
\\\\

\textbf{Summary A:} \{summary\_A\}
\\\\
\textbf{Summary B:} \{summary\_B\}
\\\\
Generate Your assessment in the following format:
\\
\textbf{Better Summary Index: [Index]}(Use A for select Summary A and B for select Summary B)

\\
\bottomrule
    \end{tabular}
    \caption{The prompt for pairwise summary quality comparison (criteria: natural)}
    \label{tab:prompt_pairwise_natural}
\end{table*}

\subsection{Prompts}
We show prompts used in our experiments in Table~\ref{tab:crf_AQ} to~\ref{tab:prompt_pairwise_natural}.
\label{sec:prompt_appendix}

\end{document}